%% file: main.tex
\documentclass[10pt,twocolumn,letterpaper]{article}

\usepackage[pagenumbers]{cvpr} 
\usepackage{graphicx}
\usepackage{amsmath}
\usepackage{amssymb}
\usepackage{booktabs}

\definecolor{cvprblue}{rgb}{0.21,0.49,0.74}
\usepackage[pagebackref,breaklinks,colorlinks,allcolors=cvprblue]{hyperref}
\input{preamble}

\def\paperID{9714} 
\def\confName{CVPR}
\def\confYear{2025}

\title{Exemplar Masking for Multimodal Incremental Learning}

\author{%
  Yi-Lun Lee\textsuperscript{$\dagger$} \quad Chen-Yu Lee\textsuperscript{$\ddagger$} \quad Wei-Chen Chiu\textsuperscript{$\dagger$} \quad Yi-Hsuan Tsai\textsuperscript{$\S$} \\
  \textsuperscript{$\dagger$}National Yang Ming
Chiao Tung University \quad \textsuperscript{$\ddagger$}Google\quad 
\textsuperscript{$\S$}Atmanity\\  
  \small{\texttt{\{yllee10727, walon\}@cs.nctu.edu.tw}}, \;  
  \small{\texttt{chenyulee@google.com}},\;  
  \small{\texttt{yhtsai@atmanity.io}}
}

\begin{document}
\maketitle
\input{sec/0_abstract}    
\input{sec/1_intro}
\input{sec/2_related}

\input{sec/3_method}

\input{sec/4_experiment}

\input{sec/5_conclusion}
{
    \small
    \bibliographystyle{ieeenat_fullname}
    \bibliography{main}
}

\input{sec/0_supp}

\input{sec/1_vis_instructBLIP}
\input{sec/1_vis_Food101}


\end{document}

%% file: preamble.tex
\usepackage[dvipsnames]{xcolor}
\usepackage{soul}
\usepackage{url}
\usepackage[utf8]{inputenc}
\usepackage{graphicx}
\usepackage{amsmath}
\usepackage{booktabs}
\usepackage{epsfig}
\usepackage{amssymb}
\usepackage{hhline}
\usepackage{multirow}
\usepackage{array}
\usepackage{wrapfig}
\usepackage{tabu}
\usepackage{floatrow, comment, enumerate}
\usepackage{color}
\usepackage{algorithmic}
\usepackage[linesnumbered,ruled,vlined]{algorithm2e}
\usepackage{subcaption}
\usepackage[utf8]{inputenc}
\usepackage{paralist}

\DeclareUnicodeCharacter{2212}{-}

\let\oldnl\nl
\newcommand{\nonl}{\renewcommand{\nl}{\let\nl\oldnl}}

\newcommand{\alan}[1]{\textcolor{black}{#1}}

\newcommand{\red}[1]{{\color{red}#1}}
\newcommand{\gray}[1]{{\color{gray}#1}}
\newcommand{\blue}[1]{{\color{blue}#1}}

\newcommand{\obj}[1]{{\colorbox{yellow}{#1}}}
\newcommand{\bg}[1]{{\colorbox{Melon}{#1}}}

%% file: sec/0_abstract.tex
\begin{abstract}
%
Multimodal incremental learning needs to digest the information from multiple modalities while concurrently learning new knowledge without forgetting the previously learned information.
There are numerous challenges for this task, mainly including the larger storage size of multimodal data in exemplar-based methods and the computational requirement of finetuning on huge multimodal models.
%
In this paper, we leverage the parameter-efficient tuning scheme to reduce the burden of fine-tuning and propose the exemplar masking framework to efficiently replay old knowledge.
Specifically, the non-important tokens are masked based on the attention weights and the correlation across different modalities, significantly reducing the storage size of an exemplar and consequently saving more exemplars under the same memory buffer. 
%
Moreover, we design a multimodal data augmentation technique to diversify exemplars for replaying prior knowledge.
%
In experiments, we not only evaluate our method in existing multimodal datasets but also extend the ImageNet-R dataset to a multimodal dataset as a real-world application, where captions are generated by querying multimodal large language models (e.g., InstructBLIP).
%
Extensive experiments show that our exemplar masking framework is more efficient and robust to catastrophic forgetting under the same limited memory buffer.
Code is available at \href{https://github.com/YiLunLee/Exemplar_Masking_MCIL}{https://github.com/YiLunLee/Exemplar\_Masking\_MCIL}.
\end{abstract}

%% file: sec/1_intro.tex
\section{Introduction}
\begin{figure}
    \centering
    \includegraphics[width=\linewidth]{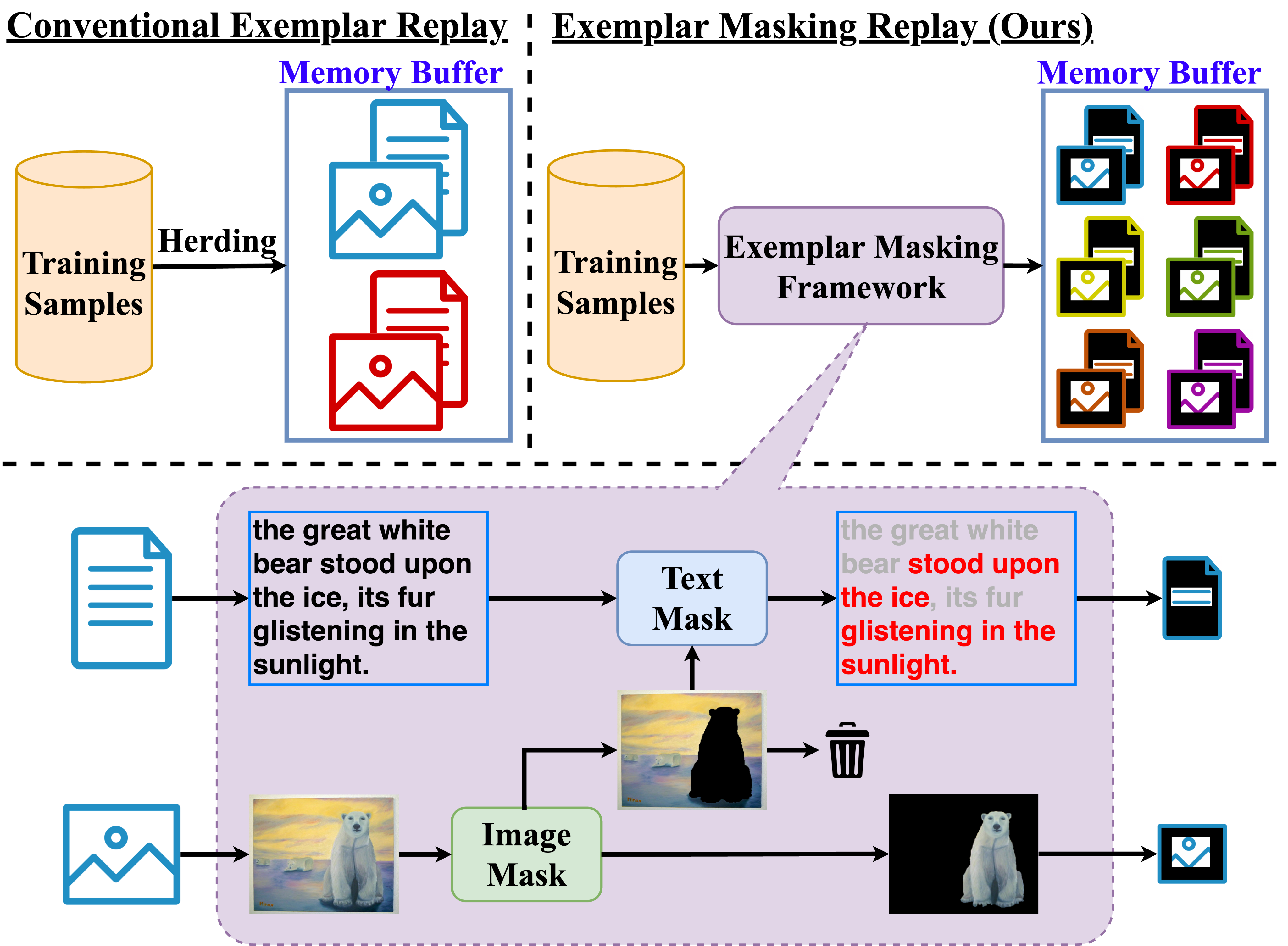}
    \caption{Illustration of exemplar replay for a new class ``ice bear''. In the conventional exemplar replay framework, only very few data samples can be stored in the limited memory buffer due to the high storage demand.
    In contrast, our exemplar masking framework preserves the important regions of the image and discards the non-important ones to reduce the storage space. Moreover, we propose to preserve the information of discarded regions via another modality (i.e., text) to retain as much information as possible. Under the same memory buffer, our framework can store more samples, contributing to more effective knowledge replay. \vspace{-1em}
    }
    \label{fig:teasor}
\end{figure}
Human perception of the real-world environment acquires knowledge from multiple senses in a continual and sequential manner. Recently, multimodal transformers \cite{kim2021vilt, lee2023multimodal, ma2022multimodal, pham2019found, wang2020transmodality, kamath2021mdetr} pre-trained on large-scale datasets demonstrate promising performance across a range of multimodal tasks, including visual recognition \cite{lee2023multimodal, ma2022multimodal}, object detection \cite{kamath2021mdetr}, multimodal sentiment analysis \cite{pham2019found, wang2020transmodality}, etc.
However, when facing the scenario that the model keeps updated with the new data coming in sequentially, such strong models suffer from the issue of catastrophic forgetting which may lead to severe performance drop of the old knowledge.
Moreover, with the growth in the size of multimodal models, finetuning the entire model with multimodal data becomes increasingly impractical under the limited computation resources. To this end, we delve into the realm of multimodal class-incremental learning (MCIL), which is a more practical learning scenario for AI agents.

In this paper, we tackle two practical challenges: 1) heavy multimodal model fine-tuning, and 2) catastrophic forgetting when learning with new data.
First, nowadays it is common to use multimodal models with billions of parameters~\cite{brown2020language,raffel2020exploring,rae2021scaling} as the pre-trained models and transfer it on the downstream tasks with finetuning. 
%
However, as the multimodal model size explosively increases, fine-tuning the entire models when new data comes inevitably results in a heavy computation cost, which is even no longer applicable under the limited computation resources. 

%
To this end, we adopt parameter-efficient tuning (PET) \cite{zaken2022bitfit, lester2021power, li2021prefix, lian2022ssf, hu2021lora} in our MCIL framework to achieve the model training efficiency while reducing the negative fine-tuning effect in retaining the old knowledge.

%
Second, to further alleviate catastrophic forgetting, exemplar replay proposed by iCaRL~\cite{rebuffi2017icarl} has emerged as a widely-used technique in incremental learning.
The core concept is to retain a limited number of past samples per class as exemplars, facilitating the retention of previous knowledge when learning new classes.
For our multimodal setup, as the volume of multimodal data increases, the capacity of storing complete exemplars in the memory buffer decreases, which makes it more difficult to keep the old knowledge.
To address this challenge and select more representative exemplars within the same buffer size, a few data-efficient methods~\cite{wang2022memory,luo2023cim} are proposed.
However, these approaches still retain redundant information, e.g., non-essential regions in images, resulting in substantial storage overhead.
In addition, for each type of data in multimodal tasks, the method to reduce redundant information may vary accordingly, which is not addressed in the prior work.

%
In this paper, we design an MCIL framework by accounting the property of each data type, thereby optimizing the space to store exemplars dynamically for each data with different modalities.
Specifically, we propose an exemplar masking method to select important tokens of exemplars according to their attention weights, so only a portion of information in representative exemplars for each class is needed to store (see Figure~\ref{fig:teasor}).
Our motivation is that, the discriminative tokens with higher attention weights may be more valuable to be preserved for replaying old knowledge, while the masked tokens containing non-discriminative information can be removed.
Moreover, to replay the masked exemplars more effectively, we further interchange the data of different modalities in the same class as a data augmentation strategy to improve data diversity.

In experiments, we not only validate our proposed method in existing multimodal datasets but also create a multimodal dataset extended from the existing ImageNet-R~\cite{hendrycks2021many} dataset in a practical manner for the MCIL task.
We generate captions for the images with multimodal large lanaguage models (MLLMs)~\cite{instructblip}. We conduct extensive experiments to explore different masking and selection strategies to demonstrate that the proposed method is able to achieve efficient and effective MCIL against other baseline approaches.
Our contributions are summarized as follows:
\begin{compactitem}
    \item We explore multimodal incremental learning in both data-efficient and memory-efficient manners, which is practical for AI agents.
    \item We propose an exemplar masking method for the exemplar-based incremental learning framework, which highly reduces the storage space of multimodal samples via adaptive masking, and thus more samples are able to be saved in the same memory buffer size.
    \item To replay the multimodal masked exemplars more efficiently, we further propose multimodal data augmentation to enrich the old exemplars, encouraging models to replay the old knowledge effectively.
    \item We extend the image classification dataset to a multimodal one by generating rich captions via MLLMs.
\end{compactitem}

%% file: sec/2_related.tex
\section{Related Work}
\subsection{Incremental Learning} 
The ability to learn new concepts without forgetting the previously acquired knowledge is essential for AI agents. There are three groups of methods for alleviating the forgetting issue.
\textbf{Parameter regularization methods} \cite{li2017lwf, Kirkpatrick3521} estimate the discrepancy between new and old models, and then adopt corresponding penalization terms to the objectives.
\textbf{Model-based methods} \cite{wang2022foster, yan2021dynamically, abati2020conditional, liu2021adaptive} preserve the model parameters for learning new classes to prevent overwriting the learned weights for previous classes.
\textbf{Replay-based methods}, as a long-lasting and widely-used method, assume there exists a limited memory buffer to store a few old samples (named as exemplars) for replaying previous knowledge. iCaRL \cite{rebuffi2017icarl} first introduces this paradigm for class-incremental learning, motivating various works to improve the performance with a limited replay buffer. \cite{wu2019large, hou2019learning, zhao2020maintaining} aim at mitigating the issue of biased classifiers, which arises due to the substantial imbalance between the number of old exemplars and newly acquired samples.
In addition, some approaches share the same high-level idea as exemplars but with different storing types, including topology-based \cite{tao2020topology}, feature-based \cite{iscen2020memory}, GAN-based \cite{shin2017continual, wu2018incremental}, and prompt-based \cite{wang2022learning, wang2022dualprompt} methods.

\begin{figure*}[!t]
    \centering
    \includegraphics[width=1\linewidth]{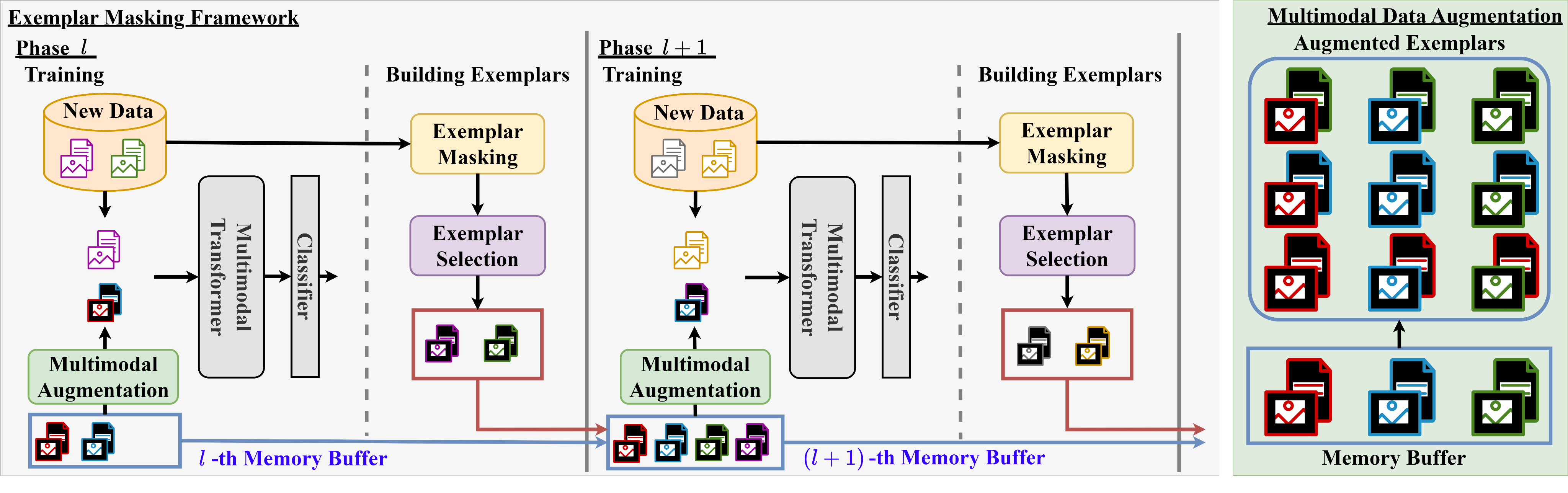}
    \caption{Overview of the proposed exemplar masking framework for multimodal class-incremental learning, including exemplar masking, exemplar selection, and multimodal data augmentation. In the $l$-th incremental phase, we first apply multimodal data augmentation on the $l$-th memory buffer and train the model with both new data and augmented exemplars. After training, we generate the masked exemplars from new data via the proposed exemplar masking and exemplar selection methods, then combine them with the memory buffer. \vspace{-1.5em}
    }
\label{fig:exemplar_masking_replay_framework_multimodal_data_augmentation}
\end{figure*}

Recently, \cite{liu2020mnemonics, wang2022memory, luo2023cim, lee2023smc, qiang2023mixture} have been proposed to improve the memory efficiency of the replay-based methods. 
Mnemonics \cite{liu2020mnemonics} distill the current new training samples into exemplars via a bi-level optimization to store more representative samples with the same quantity. 
On the other hand, numerous works \cite{lee2023smc, qiang2023mixture} focus on augmenting the exemplars with augmentation methods like mixup \cite{zhang2017mixup} to diversify the exemplars and avoid the over-memorizing issue.
Furthermore, considering the trade-off between quality and quantity, MRDC \cite{wang2022memory} adopts  JPEG compression codec to compress the samples into more compact data, while CIM \cite{luo2023cim} proposes class-incremental masking based on the class activation map (CAM) to select the discriminative regions and downsample other regions.
However, these downasmapled and non-discriminative regions sometimes do not provide useful information for replaying old knowledge, leading to waste the storage space. Also, these image compression and downsampling methods cannot be generalized to different modality data for the multimodal incremental learning scenario.
In our work, we adopt the multimodal transformer as the backbone where the multimodal data is embedded in tokens, and we propose an exemplar masking framework to discard the non-important tokens according to the attention map.



\subsection{Parameter-efficient tuning} 
As the explosion of the parameters in huge pre-trained multimodal models, finetuning the entire model on downstream tasks is not available under the limited computation resources. 
To alleviate the burden of finetuning entire models without hurting the performance, there are various parameter-efficient tuning (PET) methods to alleviate this issue. As the pioneer, \cite{houlsby2019parameter} proposes to transfer the pre-trained models on downstream tasks with lightweight adapter modules.
Prompt tuning  \cite{lester2021power} and prefix tuning  \cite{li2021prefix} propose to insert the learnable tokens (named prompts) into the input sequence in order to instruct the pre-trained models performing the downstream tasks.
Bitfix~\cite{zaken2022bitfit} adapts pre-trained models on small datasets with only training the bias-terms in the layers.
SSF~\cite{lian2022ssf} modifies the features via a linear transformation with learnable scale and shift features for transferring to new data.
LoRA~\cite{hu2021lora} injects trainable rank decomposition matrices into each transformer block to learn the adaptation on downstream tasks.
Compared to finetuning entire models, these PET methods reach a competitive or even better performance with very few learnable parameters (i.e., $< 1\%$ total model parameters). In our work, we adopt the SSF~\cite{lian2022ssf} as our PET method for efficient multimodal incremental learning in the practical scenario.

%% file: sec/3_method.tex
\vspace{-0.5em}
\section{Proposed Method}
\begin{figure*}[!t]
    \centering
    \includegraphics[width=1\linewidth]{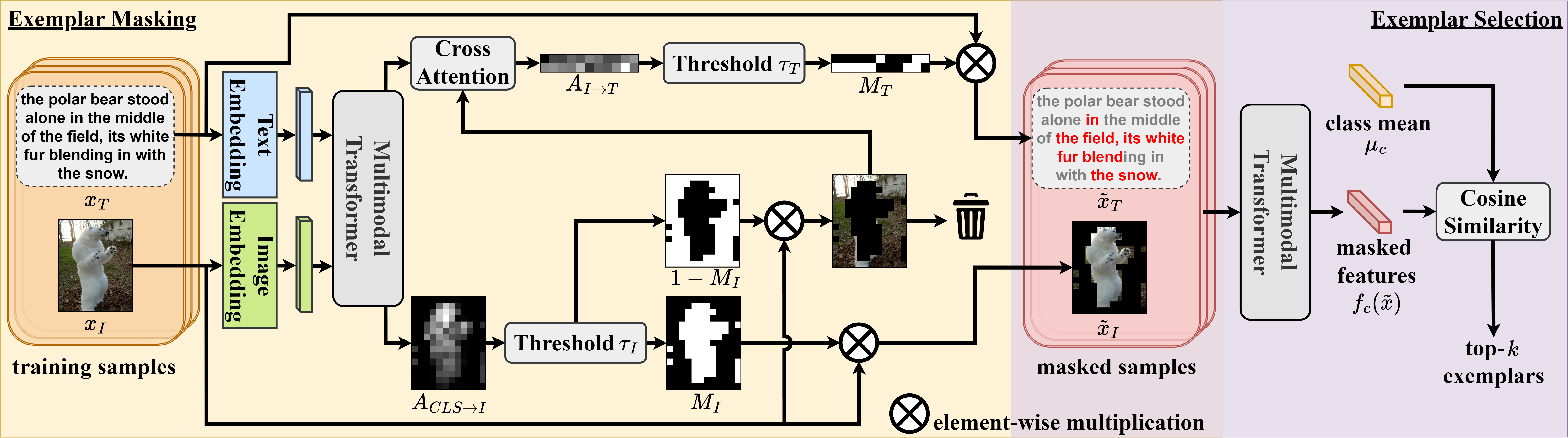}
    \caption{The overview of our proposed exemplar masking and exemplar selection methods. Given a training sample $(x_T, x_I)$ of class $c$, we first calculate the attention of the class token for the image modality $A_{CLS \to I}$ and obtain the image mask $M_I$ according to the threshold $\tau_I$. Then the masked image regions are preserved in the masked image $\Tilde{x_I}$ while the others are discarded.
    To preserve contextual information from the discarded image regions, we calculate the cross attention $A_{I \to T}$ between discarded image tokens and text tokens to obtain the text mask $M_T$ via the threshold $\tau_T$. Hence the masked text $\Tilde{x_T}$ is produced via applying text mask $M_T$ on $x_T$.
    Finally, we compute the cosine similarity between the feature $f_c(\Tilde{x})$ of the masked sample $(\tilde{x}_T, \tilde{x}_I)$ and the mean $\mu_c$ of the class $c$, in which the samples with the top-$k$ highest similarity are selected as the exemplars of class $c$ and preserved without exceeding the memory size. \vspace{-1em}
    }
    \label{fig:model}
\end{figure*}

In this paper, we focus on multimodal class-incremental learning based on the exemplar replay framework and propose an exemplar masking framework for it, as shown in Figure~\ref{fig:exemplar_masking_replay_framework_multimodal_data_augmentation}.
Without loss of generality, we consider that the multimodal data is composed of two modalities: text $T$ and image $I$. As the typical scenario of incremental learning, the data of different classes arrive sequentially in each phase, in which there are in total $L$ incremental phases. Thus the entire multimodal incremental dataset is denoted as $D = \{D^1, D^2, ..., D^L\}$ where $D^l$ indicates the data samples arriving in $l$-th phase. Basically, model in the $l$-th incremental phase is expected to learn the new classes in $D^{\text{new}} = D^l$ while retaining the previous knowledge of the old classes in $D^{\text{old}} = \{D^1, D^2, ..., D^{l-1}\}$. Due to limited size of the memory buffer, we cannot access the entire old training samples $D^{\text{old}}$, and thus we only preserve a few representative samples $D^{\text{exp}}$ (named as \textit{exemplars}) from previous data (i.e., $D^{\text{exp}} \subset D^{\text{old}}$) via the herding algorithm \cite{rebuffi2017icarl}. Overall, the model is trained with available training samples in $D^{\text{new}} \cup D^{\text{exp}}$ during the $l$-th incremental phase. 

We adopt the multimodal transformer ViLT~\cite{kim2021vilt} as our backbone model, which is pre-trained on multiple large image-text datasets. 
Instead of fine-tuning the entire model, we adopt parameter-efficient tuning (PET) methods to mitigate computational demands, where we utilize the SSF \cite{lian2022ssf} as our main PET scheme in this paper.

\subsection{Exemplar Masking for Replay}
\label{sec:EMR}
The typical manner of keeping several original/complete exemplars of old classes could lead to fewer stored samples when dealing with multimodal data due to limited memory buffer.
Therefore, it is crucial to maximize the storage efficiency of the memory buffer such that it only preserves useful information of old data for replaying purposes. To this end, for each exemplar (which is composed of text $x_T$, image $x_I$, and the ground truth class label $y$) to be kept in the buffer, we propose to only store its essential tokens while discarding the non-important ones according to the attention weight of the class token $x_\text{CLS}$\footnote{We note that our backbone is ViLT, where the input exemplar is tokenized through the transformer blocks to have the resultant class token.}, as shown in Figure~\ref{fig:model}.
To be detailed, given the attention map $A_{\text{CLS}}$ of the class token $x_{\text{CLS}}$, we separate it into $A_{CLS \to T}$ and $A_{CLS \to I}$ for the text and image modalities respectively, followed by using them to create the masks for masking the text $x_T$ and image $x_I$. As the image modality requires a larger storage space, we first consider masking the image tokens.

\noindent\textbf{Masking image exemplars.}
Before moving to the phase $l+1$, the model has learned to recognize samples of the new classes $C^{\text{new}(l)}$ added in the phase $l$, in which some selected exemplars of classes $C^{\text{new}(l)}$ will be replayed in the next phase for retaining the knowledge.
Therefore, the attention maps produced by the model are expected to focus more on the image content related to classes $C^{\text{new}(l)}$, meaning that the image tokens with higher attention weights involve the most information about $C^{\text{new}(l)}$.
To effectively mask out the non-discriminative tokens and preserve the informative tokens, we find that the mean of the image attention map $A_{CLS \to I}$ serves as a suitable threshold $\tau_I$ for producing the mask of image modality $M_I =  \{ m_I(i) \mid i = 0, 1, ..., N_I \}$, where $N_I$ is the number of image tokens:
\vspace{-1em}
\begin{align}
 \tau_I &= \frac{1}{N_I} \sum_{i=0}^{N_I} A_{CLS \to I}(i),\label{equ:image_threshold} \\
 m_I(i) &= \left\{\begin{matrix} 
1, & A_{CLS \to I}(i) \geq  \tau_I, \\ 
0, & \text{otherwise}.
\end{matrix} \right.
\label{equ:image_mask} 
\end{align}


\alan{ 
Setting the threshold for masking to the mean of attention weights rather than a fixed value allows for dynamic adjustment based on input content. This adaptive threshold enhances flexibility of our method across various datasets.
}

\noindent\textbf{Masking text exemplars.} 
Although some image regions are discarded due to low attention weights (i.e., being less important in the image domain), they may still provide \textit{contextual information} which is beneficial for recognizing the target classes (e.g., the background with an ``ice lake'' could be helpful to recognize the class ``ice bear''). Hence, we would still like to maintain the contextual information hidden behind masked image tokens in our replay buffer.

To this end, we instead store such information via a memory-efficient manner by leveraging the corresponding text tokens.
As the second modality, text tokens are encouraged to preserve not only the text-related information but also the discarded information from the image modality according to the attention weights across two modalities.
To be specific, we first calculate cross-attention weights $A_{I \to T}(i)$ of text tokens with respect to image tokens (i.e., the attention from image tokens to text tokens), especially on the attention of masked image tokens (i.e., with $m_I(i) = 0$). Then we compute the threshold $\tau_T$ as follows for extracting the text tokens that are more relevant to the masked image tokens, i.e., producing the mask of text modality $M_T =  \{ m_T(i) \mid i = 0, 1, ..., N_T \}$:
\begin{align}
 \tau_T =& \frac{1}{N_T} \sum_{i=0}^{N_T} \frac{1}{N_{m_I=0}} \sum_{j=0}^{N_I} \mathbb{I}(m_I(j)=0)A_{I \to T}(j,i), \label{equ:text_threshold}\\
 m_T(i) &= \left\{\begin{matrix}
1, & A_{I \to T}(i) \geq  \tau_T, \\ 
0, & \text{otherwise}.
\end{matrix}\right.
\label{equ:text_mask} 
\end{align}
where $\mathbb{I}(\cdot)$ is an indicator function and $A_{I \to T}(j,i)$ is the cross-attention value of text token $i$ on the image token $j$, while $N_T$ and $N_{m_I=0}$ are the number of the text tokens and the masked image tokens respectively.

Eventually, the multimodal exemplar after applying our masking strategy as masked exemplars becomes: 
\begin{equation}
\Tilde{x}_T=M_T\otimes x_T, \quad \Tilde{x}_I=M_I\otimes x_I,
\label{equ:masked_data} 
\end{equation}
where $\otimes$ is the element-wise multiplication. %
\alan{
Please note that we leverage the attention map from the backbone model instead of applying additional models to identify the important regions to be preserved.
}

\noindent\textbf{Selection of masked exemplars.}
In the typical setting of replay-based incremental-learning methods, the herding selection strategy is adopted to select $k$ exemplars in which they are the most representative samples of the target class according to the class center/mean.
Specifically, the first few samples being closest to the class center are selected based on the feature space extracted by the model.
%
In the proposed method, we adopt the herding strategy used in iCaRL~\cite{rebuffi2017icarl} as well, but we use the cosine similarity for measuring the distance between the class mean and masked training samples (i.e., the samples after applying our masking operation), in which the masked training samples with the top-$k$ highest similarity to their respective class mean are selected as exemplars (see the right side of Figure~\ref{fig:model}).

\subsection{Multimodal Data Augmentation}
Although our exemplar masking method helps store more exemplars for the old/learned classes in the replay buffer of fixed size, there still exists the imbalance issue between the old and new classes. That is, the amount of exemplars used to replay the knowledge of old classes is much less than the training samples of new classes. Moreover, using the same exemplars for replaying old classes in different incremental phases would cause the over-memorizing issue (i.e., the model only memorizes the exemplars and does not generalize to other data samples of the same class), as described in \cite{lee2023smc}. To this end, we propose a simple yet effective data augmentation technique upon the exemplars to enrich the training samples of replaying old classes, thus alleviating both class-imbalance and over-memorizing issues. 

As mentioned previously, for a multimodal data sample, the masked image mainly represents the object of the corresponding class, while the masked text provides complementary and auxiliary information. Hence, our multimodal data augmentation interchanges either the images or text descriptions between an arbitrary pair of exemplars from the same class (see the right side of  Figure~\ref{fig:exemplar_masking_replay_framework_multimodal_data_augmentation}). This creates numerous diverse training samples to replay the old knowledge.

\subsection{Training Objective}
In the incremental phase $l$, we would have the masked exemplars $\tilde{D}^{\text{exp}}$ of old classes $C^{\text{old}(l)}$ that are seen from previous phases, as well as the training samples $D^{\text{new}}$ of new classes $C^{\text{new}(l)}$ arriving at phase $l$. Our objective function for multimodal class-incremental learning is based on the cross-entropy function, where its optimization in the phase $l$ is driven by the available training data $\tilde{D}^{\text{exp}} \cup D^{\text{new}}$.
Here, each training sample is composed of the multimodal data $\mathbf{x}$ (i.e., image and text) and the corresponding class label $y$.
The $\mathcal{L}_{\text{CE}}$ loss for every training sample is defined as: 
\begin{equation}
\label{equ:total_objective}
\begin{aligned}
    \mathcal{L}_{\text{CE}}(\mathbf{x},y) = \sum_{c=0}^{\mathcal{C}^l} -\delta_{c=y} log(p_c(\mathbf{x}),c),
\end{aligned}
\end{equation}
where $\mathcal{C}^l = \left|C^{\text{old}(l)}\right|+\left|C^{\text{new}(l)}\right|$ denotes the total number of classes in the current incremental phase $l$ and $p_c(\mathbf{x})$ is the output prediction on the class $c$. Please note that, in order to avoid the catastrophic forgetting caused by overriding the previously learned classifier weights, we adopt the masked logit trick~\cite{wang2022learning, wang2022dualprompt}. This operation masks the logits of old classes for training samples of new classes, while masking the logits of new classes for exemplars of old classes.

%% file: sec/4_experiment.tex
\section{Experimental Results}
\noindent\textbf{Dataset.} 
The evaluation of multimodal class-incremental learning is based on the well-known multimodal classification dataset UPMC Food-101~\cite{wang2015food101} and a newly proposed dataset proposed (denoted as \textbf{MM-ImageNet-R})
, which is stemmed from the ImageNet-R \cite{hendrycks2021many} but has the carefully-designed multimodal extension (i.e., extended from only having image modality to having both text and image modalities). 
Specifically, given an image from ImageNet-R, we derive its text description automatically by querying the multimodal large language model, InstructBLIP\cite{instructblip}. with the following prompts: ``What’s in this image? Please use 100 words to describe the image content in detail.''

\noindent\textbf{Incremental setting.}
We follow the common incremental setting in \cite{rebuffi2017icarl}, where the dataset is equally split into $L$ subsets without overlapping classes for $L$ incremental phases. The default memory buffer size is set to support 5 raw multimodal samples per class. We adopt the average incremental learning accuracy $\bar A = \sum_{l=1}^{L} A^i$ as our evaluation metric, where $A^l$ is the accuracy of all the seen classes at the end of $l$-th incremental phase.
We will make the datasets, source codes, and models available to the public.

\subsection{Quantitative Results}
In our experiments, we mainly focus on two critical properties: 1) the data efficiency of the stored exemplars, and 2) the parameter-efficient methods for multimodal incremental learning. The compared baselines include: 1) finetuning the entire model (denoted as \textbf{FT}) in each incremental phase, and 2) the parameter-efficient tuning (PET) method (i.e., \textbf{SSF} \cite{lian2022ssf} is adopted in experiments) that only learns the additional few parameters in each incremental phase. Both of them are based on the conventional exemplar-replay framework.
Moreover, in order to compare with other data-efficient exemplar-replay methods on the scenario of multimodal incremental learning, a baseline is built by following the concept of \textbf{CIM}~\cite{luo2023cim} which selects the important image regions according to the CAM map and downsamples the non-discriminative image regions by 4 times (i.e., half on both width and length) to compress the size of exemplars.
\begin{table*}[!ht]
    \centering
    \caption{Average incremental accuracy on our MM-ImageNet-R and UPMC Food-101 datasets with different numbers of incremental phases $L$=5, 10, 20. The memory size is 5 raw multimodal samples/class. \textbf{Ours} stands for our overall exemplar masking framework, while 
    \textbf{MDA} stands for multimodal data augmentation. \vspace{-1em}
    }
    \label{table:quantitative}
    \resizebox{0.8\linewidth}!{
    \begin{tabular}{ccccccccc}
    \hline
    \multirow{2}{*}{Methods}  &
    \multirow{2}{*}{\shortstack{\# of \\ Parameters}} &
    \multicolumn{3}{c}{
    MM-ImageNet-R} &&
    \multicolumn{3}{c}{
    UPMC Food-101} \\ \cline{3-5} \cline{7-9}
    & & $L$=5 & $L$=10 & $L$=20 && $L$=5 & $L$=10 & $L$=20 \\
    \hline 
    FT  & \multirow{4}{*}{112M} & 79.72 & 77.31 & 75.22  && 86.43 & 81.61 & 76.97 \\
    FT + CIM~\cite{luo2023cim}  & & 80.33 & 78.41 & 76.46 && 87.15 & 82.68 & 78.98 \\
    FT + Ours (w/o MDA)  & & 82.06 & 79.39 & 76.23 && 88.03 & 84.54 & 79.49 \\
    FT + Ours  & & \textbf{82.97} & \textbf{80.72} & \textbf{78.19} && \textbf{88.08} & \textbf{85.91} & \textbf{80.53} \\
    \hline 
    SSF  & \multirow{4}{*}{206K} & 80.58 & 77.03 & 75.26 && 86.66 & 81.62 & 76.91 \\
    SSF + CIM~\cite{luo2023cim}  & & 80.97 & 78.21 & 77.23 && 86.39 & 83.48 & 79.25 \\
    SSF + Ours (w/o MDA)  & & 81.80 & 79.05 & 77.14 && \textbf{87.20} & 83.03 & 79.79 \\
    SSF + Ours  & & \textbf{82.76} & \textbf{80.55} & \textbf{78.64} && 86.73 & \textbf{84.51} & \textbf{80.68}
    \end{tabular}
    }
\end{table*}

We summarize quantitative results of our proposed methods on MM-ImageNet-R and UPMC Food-101 in Table~\ref{table:quantitative}. We draw several observations:\\
\indent 1) SSF reaches a comparable performance with FT and the required parameters for training are less than 0.3\% of total model parameters, showing that SSF (i.e., a parameter-efficient tuning method) is more applicable with limited computation resources. \\
\indent 2) In comparison to baselines, our exemplar masking framework without multimodal data augmentation (denoted as \textbf{Ours w/o MDA}) improves performance as the replaying is benefited from having more exemplars thus becomes more effective. In addition, with the multimodal data augmentation  (denoted as \textbf{Ours}), the performance is further boosted to achieve the best across two training schemes (i.e., by FT or SST).\\
\indent 3) When the number of incremental phases $L$ increases, the model variants adopting our multimodal data augmentation improve more, e.g., in Table~\ref{table:quantitative}, ``FT + Ours'' improves ``FT + Ours (w/o MDA)'' by 1.96\% for $L$=20 on MM-ImageNet-R, showing that our model is less sensitive to the forgetting issue under the long-term incremental case, with the help from more diverse exemplars generated by our multimodal data augmentation.\\ 
\indent 4) Although applying CIM to baselines also improves the performance, the gain is not optimal when compared with ours. The reason is that the non-important regions are preserved in low-resolution and still require additional storage space, leading to inefficient memory usage.
In contrast, our exemplar masking framework discards non-important regions but still preserves the contextual information of discarded regions in another modality (i.e., text), which makes storing exemplars more efficient and replaying exemplars more effective.

\begin{figure*}[!t]
    \centering
    \LARGE
    \resizebox{1.0\linewidth}!{
        \begin{tabular}{cccccccccc}
         (a)
        &\begin{tabular}{c}\includegraphics[width=0.3\textwidth]{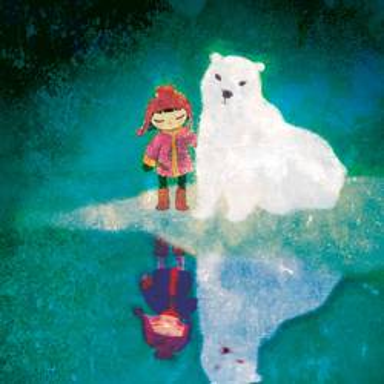}\end{tabular}
        &\begin{tabular}{c}\includegraphics[width=0.3\textwidth]{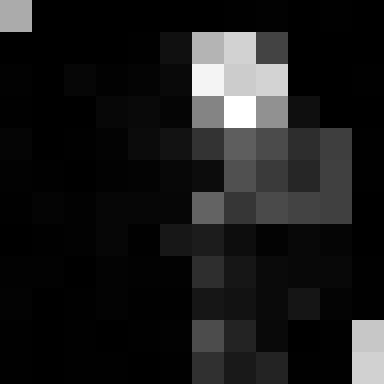}\end{tabular}
        &\begin{tabular}{c}\includegraphics[width=0.3\textwidth]{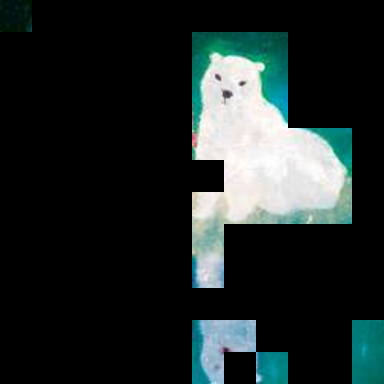}\end{tabular}
        &\multicolumn{6}{m{20cm}}{ [CLS] \gray{the \red{image} displays an \red{animation} - style character art, with \red{a} young \red{\bg{girl}} standing next to an \red{antler}ed \red{\obj{animal}}. \red{it is likely a \obj{polar bear}}, but its features appear \red{distorted} in the painting, giving a \red{surreal}istic feel to the piece. a \red{\bg{girl}} stands close to the polar bear, looking intently \red{at} it. \red{in} the background, there\red{ is a \bg{snowy ice landscape}} or \red{a} reflection\red{ of the \bg{girl}} on the \red{\bg{water}}. the painting \red{is designed} as a \red{children}'s \red{book} cover and could possibly be used as a coloring page.} [SEP]}
        \\               
        (b)
        &\begin{tabular}{c}\includegraphics[width=0.3\textwidth]{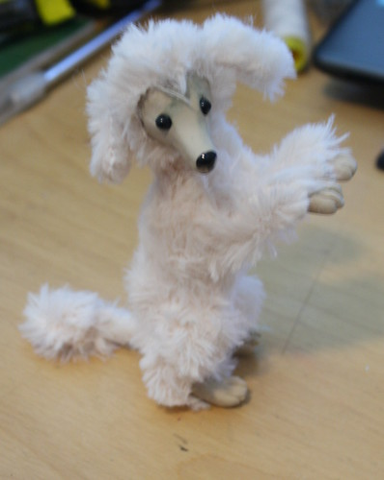}\end{tabular}
        &\begin{tabular}{c}\includegraphics[width=0.3\textwidth]{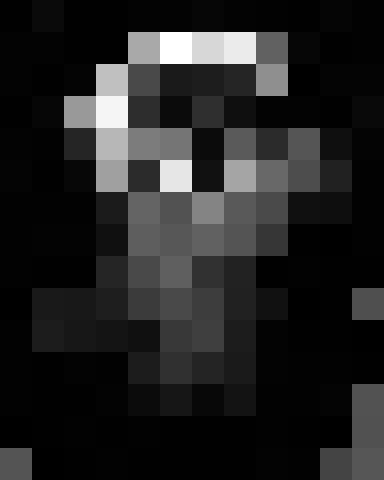}\end{tabular}
        &\begin{tabular}{c}\includegraphics[width=0.3\textwidth]{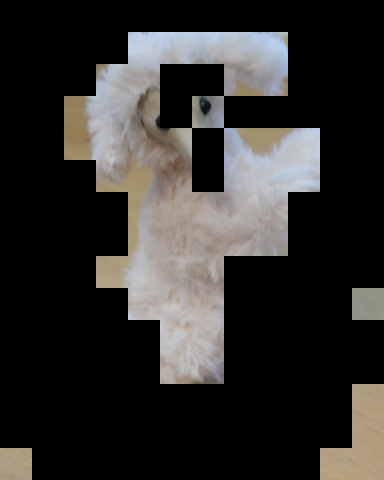}\end{tabular}
        &\multicolumn{6}{m{20cm}}{ [CLS] \gray{\red{the image depicts a \obj{small white dog}}, \red{likely} dressed up in a \red{\obj{plush} costume or }outfit featuring a white \red{\obj{fur}} texture, standing on a \red{\bg{hardwood table}. the} toy \red{\obj{dog}}, which \red{appears} to be \red{realistic} and cute, \red{is} sitting in a particular position, most \red{\obj{likely playing}} with \red{or \obj{interacting}} with some other object or object. the image also \red{includes} a hand, seen on the \red{right} side of the \red{\bg{table}}, looking as \red{if} it \red{might be interacting} with \red{the} dog. apart \red{from} the \red{\obj{dog}} and hand, there’s an \red{unidentified object} sitting on the edge of the \red{\bg{table}}, \red{possibly} nearby \red{the \obj{dog}}.} [SEP]}
        \\  
        (c)
        &\begin{tabular}{c}\includegraphics[width=0.3\textwidth]{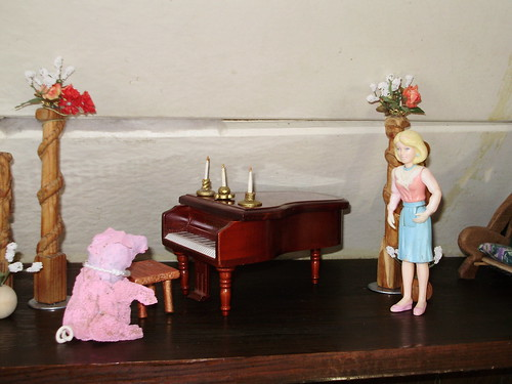}\end{tabular}
        &\begin{tabular}{c}\includegraphics[width=0.3\textwidth]{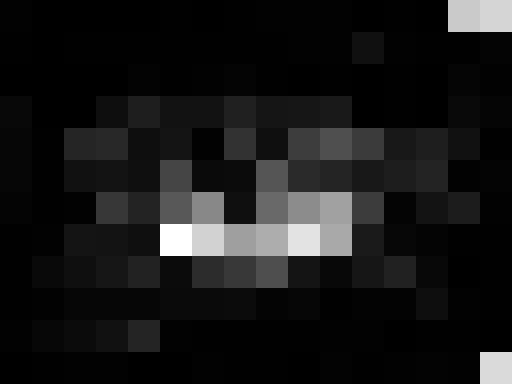}\end{tabular}
        &\begin{tabular}{c}\includegraphics[width=0.3\textwidth]{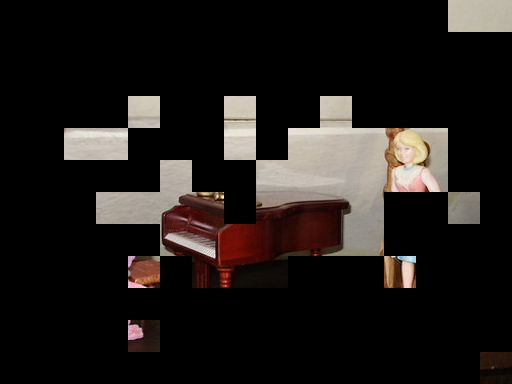}\end{tabular}
        &\multicolumn{6}{m{20cm}}{ [CLS] \gray{\red{this image features a \bg{wooden shelf displaying}} several small \red{\bg{figurines}} of \red{\bg{dolls}} and \red{miniature} objects, including \red{\obj{a piano}}, \red{\obj{a bench}}, and a chair. the \red{ is} filled with various \red{playful} items \red{like} dolls, \red{tea} cups, and flowers. the overall display gives an impression of creativity, care, and attention to detail.} [SEP]}
        \\     
        & Input Image $x_I$ & Attention Map $A_{CLS \to I}$ & Masked Image $\Tilde{x_I}$ & \multicolumn{6}{c}{Masked Text $\Tilde{x_T}$} \vspace{-0.5em}
    
        \end{tabular}
    }
    \vspace{-0.2em}
    \caption{Examples of the masked exemplars and the corresponding attention maps from different classes, including (a) ice bear, (b) poodle, and (c) grand piano.
    We denote colors for the masked text in \red{red} and the discarded text in \gray{gray}, while also highlighting the important words related to \bg{contextual information} as well as \obj{class-related information}. \vspace{-1.7em}
    }
    \label{fig:examples}
\end{figure*}

\subsection{Qualitative Results}
In Figure~\ref{fig:examples}, we visualize the masked exemplars and the corresponding attention map. For the masked texts, we highlight the words related to the target objects (with the yellow color box) and the words related to the contextual information from the discarded regions (with the melon color box). 
As shown in Figure~\ref{fig:examples}, the attention map $A_{CLS \to I}$ focuses on the region of the target objects, and thus the important regions related to objects are selected by our proposed exemplar masking method and preserved as the masked images (e.g., faces and bodies of the target animals). Moreover, the masked texts driven by our exemplar masking method indeed preserve words related to the class object and descriptions related to the discarded regions, both of which are beneficial for the recognition of the corresponding classes. 
For example, as shown in the first row of Figure~\ref{fig:examples}, the word ``polar bear'' and ``animal'' are related to the ice bear while the word ``snowy ice langscape'', ``girl'', and ``water'' are related to the discarded image regions that indicate the environment where the bear is.

\subsection{Comparison with Other CIL Methods}
\alan{To better validate the effectiveness of our proposed method, we then provide additional comparisons with: 1) another data-efficient exemplar-based method MRDC~\cite{wang2022memory}, which adopts JPEG to compress images in order to reduce the storage size, and 2) three state-of-the-art unimodal incremental learning approaches, L2P~\cite{wang2022learning}, DualPrompt~\cite{wang2022dualprompt}, and EASE~\cite{zhou2024ease}, which also adopt pre-trained models and perform parameter-efficient fine-tuning. Note that none of the above methods are originally designed for multimodal scenarios, so we adapt their implementations to our setting and tune the training hyperparameters if needed for fair comparisons. 
As results shown in Table~\ref{table:comparison_sota_imagenetR}, we show that simply adapting unimodal methods (L2P, DualPrompt, and EASE) has suboptimal performance even compared to our SSF baseline, demonstrating that the multimodal incremental learning is a more challenging task and requires more careful designs. 
Moreover, our exemplar masking framework consistently outperforms other data-efficient algorithms (MRDC, CIL), showing that our proposed framework can efficiently store more exemplars in the limited memory buffer and thus replay exemplars more effectively.}

\begin{table}[t]
    \centering
    \caption{Average incremental accuracy on our MM-ImageNet-R dataset with different numbers of incremental phases $L$=5, 10, 20. The memory size is 5 raw multimodal samples per class. \vspace{-0.3em}
    }
    
    \label{table:comparison_sota_imagenetR}
    \scriptsize
    \resizebox{0.88\linewidth}!{
    \begin{tabular}{ccccc}
    \hline
    \multirow{2}{*}{Methods}  &
    \qquad &
    \multicolumn{3}{c}{
    MM-ImageNet-R} \\ \cline{3-5} 
    &\qquad & $L$=5 & $L$=10 & $L$=20 \\
    \hline 
    L2P~\cite{wang2022learning}  & \qquad & 68.17 & 69.21 & 66.23 \\
    DualPrompt~\cite{wang2022dualprompt}  & \qquad & 77.66 & 75.57 & 70.47 \\
    EASE~\cite{zhou2024ease}  & \qquad & 78.06 & 76.84 & 73.91 \\
    \hline 
    SSF  & \qquad & 80.58 & 77.03 & 75.26 \\
    SSF + MRDC~\cite{wang2022memory}  & \qquad & 80.77 & 78.66 & 76.13 \\
    SSF + CIM~\cite{luo2023cim}  & \qquad & 80.97 & 78.21 & 77.23 \\
    SSF + Ours  & \qquad & \textbf{82.76} & \textbf{80.55} & \textbf{78.64} 
    \end{tabular}
    }
    \vspace{-1.8em}
\end{table}

\subsection{Ablation Studies}\label{sec:ablation}
To validate our designs for the proposed exemplar masking framework, we conduct several ablation studies, including masking references, masking thresholds, and the usage of memory. 
We also discuss different masking methods in the supplementary materials.
%
We define the ``preserved ratio'' to represent the proportion of the preserved tokens after masking, and ``\# of exemplars'' indicates the actual number of masked exemplars stored in the memory buffer with the capacity of 5 raw multimodal samples per class.

\noindent\textbf{Masking references.} In this paper, exemplar masking preserves the important tokens and discards the non-important ones according to the masking reference. In addition to adopting the attention map as masking reference, we further experiment with several design choices, such as 1) \textbf{Entropy}, 2) class activation map (\textbf{CAM}), 3) \textbf{GradCAM}~\cite{selvaraju2017grad}, and 4) \textbf{Random} masking.
To be specific: 1) The entropy of attention weights for each token is calculated, where the high entropy indicates that the token attends to others more equally, while the low entropy implies that the token only focuses on specific few tokens; 2) CAM is based on the activation value from the last layer of the model, reflecting the response of each token to the given class; 3) GradCAM further considers the gradient of the target class flowing into the last layer of the model, highlighting the important tokens for predicting that class; 4) Random masking simply draws some random tokens to discard them. Note that the first three design choices follow the same procedure as our proposed method (i.e., using attention maps as the masking reference) to first compute the mean of their respective maps and threshold the maps by the resultant mean to construct the final masks.

In Table~\ref{table:mask_reference}, using the attention map in our proposed method reaches the best performance, indicating that the attention map is more suitable for reflecting the importance of input tokens. The reason is that the attention map is a more meaningful way that indicates the class token learned by accumulating the information from text and image tokens.

\begin{table}[t]
    \centering
    \caption{Ablation study of adopting different design choices as the masking reference (see the first paragraph of Section~\ref{sec:ablation} for more details). The memory size is 5 raw samples per class.\vspace{-0.5em}
    }    
    \label{table:mask_reference}    
    \resizebox{0.8\linewidth}!{
        \begin{tabular}{c|c|c|c}
        \multirow{2}{*}{\shortstack{Masking\\ Reference}}  & \multirow{2}{*}{\shortstack{Preserved\\ Ratio}} & \multirow{2}{*}{\shortstack{\# of\\ Exemplars}} & \multirow{2}{*}{$\bar A$} \\
        &&&\\
        \hline
        Attention Map  & 0.34 & 14.08 & \textbf{80.55}\\
        Entropy  & 0.48 & 9.78 & 79.89\\
        CAM & 0.38 & 11.91 & 79.36\\
        GradCAM & 0.45 & 10.68 & 79.66\\
        Random  & 0.35 & 14 & 79.58 \\ 
        \hline 
        \end{tabular}      
    }
    \vspace{-0.5em}

\end{table}

\noindent\textbf{Masking threshold.} The value of the masking threshold determines the amount of preserved tokens, which also affects the numbers of the stored exemplars directly. With the higher threshold, the preserved tokens are fewer and thus the number of exemplars increases, and vice versa. We validate our method on 5 different threshold values ($\tau_I$ for $A_{CLS \to I}$ and $\tau_T$ for $A_{I \to T}$) in Table~\ref{table:mask_threshold}. We find no obvious difference in performance when the threshold is equal to or higher than the attention mean, while the performance drops with a much lower threshold. This phenomenon illustrates our key motivation:
replaying with fewer but more important tokens per exemplar yet saving more exemplars is the better way to alleviate the forgetting issue.
\alan{
We also compare our masking threshold based on the mean of the attention maps with the fixed pre-determined threshold for image $\tau_i$ and text $\tau_t$, as shown in the last two rows in Table~\ref{table:mask_threshold}. We observe that even with a similar number of exemplars and preserved ratio, the model with a fixed masking threshold performs worse. This highlights that dynamically determining the masking threshold based on attention maps is more effective in preserving important regions given various input contents.
}

\begin{table}[!t]
    \caption{Analysis of the masking threshold (either $\tau_I$ or $\tau_T$), where $\mu$ and $\sigma$ are the mean and the standard deviation of the attention map (either $A_{CLS \to I}$ or $A_{I \to T}$). \vspace{-0.3em}
    }
    
    \centering
    \resizebox{1\linewidth}!{
        \begin{tabular}{c|c|c|c}
        \multirow{2}{*}{\shortstack{Masking\\ Threshold $\tau$}}  & \multirow{2}{*}{\shortstack{Preserved\\ Ratio}} & \multirow{2}{*}{\shortstack{\# of\\ Exemplars}} & \multirow{2}{*}{$\bar A$} \\
        &&&\\
        \hline
        $\mu + 0.5*\sigma$ & 0.21 & 21.34 & 80.24\\
        $\mu + 0.25*\sigma$ & 0.27 & 16.67 & \textbf{81.03}\\
        $\mu$ & 0.34 & 14.08 & 80.55\\
        $\mu - 0.25*\sigma$ & 0.46 & 10.18 & 80.18\\
        $\mu - 0.5*\sigma$ & 0.66 & 7.1 & 78.82\\
        \hline
        $\tau_i=0.005, \tau_t=0.001$ & 0.38 & 12.60 & 79.89\\
        $\tau_i=0.004, \tau_t=0.001$ & 0.42 & 11.26 & 79.86\\
        \hline
        \end{tabular}      
    }
    \vspace{-0.5em}
    \label{table:mask_threshold}
    
\end{table}

\begin{table}[!t]
    \caption{Comparisons under the same memory size (raw samples per class) or a similar number of exemplars. \vspace{-0.3em}
    }
    \label{table:mem_ablations}
    \centering
    \resizebox{0.75\linewidth}!{
            \begin{tabular}{c|c|c|c}
            \multirow{2}{*}{Methods} & \multirow{2}{*}{\shortstack{\# of\\ Exemplars}} & \multirow{2}{*}{\shortstack{Memory\\ Size}} & \multirow{2}{*}{$\bar A$}\\
            &&&\\
            \hline
            \multirow{2}{*}{SSF} &5 &5 & 77.03\\
             &14 &14 & 79.79\\ \hline
            SSF + CIM & 9.91 & 5 & 78.21 \\            
            SSF + Ours & 14.08 & 5 & \textbf{80.55} \\            
            \hline
            \end{tabular}       
    }
    \vspace{-1em}
\end{table}

\noindent\textbf{Discussion on memory.} In Table~\ref{table:mem_ablations}, we find that under the same memory size (i.e., 5 raw multimodal samples per class), our proposed exemplar masking framework saves more exemplars in buffer for replaying the old knowledge, thus improving the performance by a large margin. Furthermore, we observe that even with a similar number of exemplars, our method ``SSF+Ours'' with 14.08 masked exemplars per class still outperforms the ``SSF'' baseline based on conventional exemplar replay, which has 14 exemplars per class. More importantly, our memory demand for saving such amount of masked exemplars is 2.8 times smaller than the ``SSF''. This observation implies that there is no need to save the entire multimodal samples for replaying old knowledge, as most regions are redundant and do not bring significant benefits. 
\alan{
Specifically, we calculate the storage reduction for the exemplars in the first incremental phase. Given the MM-ImageNet-R dataset, the raw exemplar data occupies 3.60 MB, while the masked exemplars use only 1.29 MB, saving 64.17\% (2.31 MB) of storage.
}

\begin{figure}[!t]
    \centering
    \includegraphics[width=0.88\linewidth]{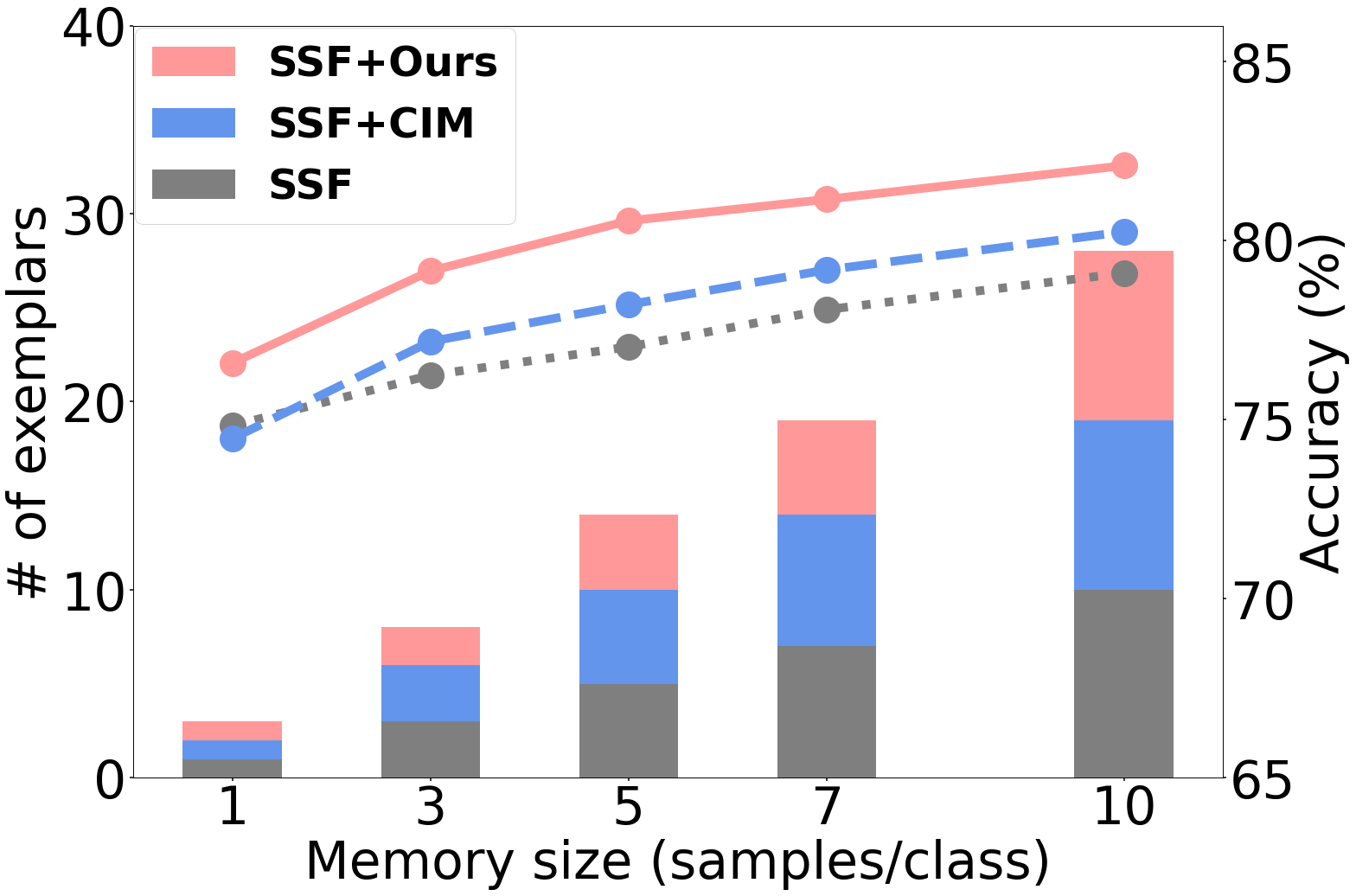}
    \vspace{-0.5em}
    \caption{Experimental results under the different constraints of the memory buffer size. Our proposed method preserves more exemplar samples under the same limited storage space and improves the baseline by a large margin. \vspace{-2em}
    }
    \label{fig:mem_ablations}
\end{figure}

Furthermore, we compare the performance of baselines with different memory buffer sizes in Figure~\ref{fig:mem_ablations}. As the CIM~\cite{luo2023cim} still requires storage space for non-discriminative tokens, the number of exemplars is limited and thus cannot reach the optimal performance. In contrast, under the same memory size, our exemplar masking framework preserves more exemplars without keeping the redundant tokens and thus outperforms baselines. 

%% file: sec/5_conclusion.tex
\vspace{-0.5em}
\section{Conclusions}
\vspace{-0.3em}
In this paper, we propose a data-efficient exemplar masking framework with the parameter-efficient tuning method for multimodal incremental learning.
Our method involves masking exemplars based on attention weights, preserving valuable discriminative tokens while discarding less important ones.
This technique allows us to store a greater number of masked exemplars within the same memory buffer. In addition, we have developed a multimodal data augmentation strategy that facilitates the exchange of multimodal data within the same class, enhancing both generalization capabilities and replay performance. Moreover, as a practical application, we extend the existing image dataset to a multimodal dataset by creating captions via captioning models and refining it with large language models. Extensive experiments and ablation studies demonstrate the efficiency and effectiveness of our framework.

%% file: sec/0_supp.tex
\clearpage
\maketitlesupplementary

\section{Ablation Study of masking methods}
\vspace{-0.5em}
\label{sec:ablation_supp}
\alan{
The design choices for the masking method involve two factors: the order of masking (i.e., which modality to be masked first) and the cross-attention strategy (i.e., what information should the masking strategy capture in the secondary modality). Specifically, there are two ways of cross-attention we investigate: whether the second modality captures 1) the contextual information from the discarded tokens (denoted as \textbf{complementary}), or 2) the information of the preserved tokens (denoted as \textbf{relevant}) of the first-masked modality.
%
\textbf{Complementary} method preserves the information from discarded regions, completing the information of entire image contents with different modalities. \textbf{Relevant} method keeps the auxiliary information related to the preserved regions, enhancing the information of target objects for recognition. 
}

\alan{In Table~\ref{table:mask_method}, we show the results of considering various designs.
In the first two rows, we observe that when masking the image first, using the different cross-attention designs for masking texts has competitive results, indicating that both designs encourage the masked texts to preserve different helpful information to assist models in replaying old knowledge. 
Particularly, the model with the complementary method reaches the best performance since it can obtain more diverse training samples via multimodal data augmentation and thus result in better replaying of exemplars. Moreover, the performance gain between the models applying these two designs increases when the number of incremental learning phases increases, validating the better effectiveness of using the complementary method for masking the second modality.
In the case of masking texts first, the masked text may not provide as much discriminative information as the masked image does, leading to suboptimal performance.
}

%
\begin{table*}[!ht]
    \caption{Ablation study of different design choices for the masking method (see the second paragraph of Section~\ref{sec:ablation_supp} for details). 
    }
    
    \label{table:mask_method}
    \centering
    \resizebox{0.7\linewidth}!{
        \begin{tabular}{c|c|c|c|c}
            \multirow{2}{*}{\shortstack{First-masked \\ Modality}} 
            & \multirow{2}{*}{\shortstack{Second-masked \\ Modality}} 
            & \multirow{2}{*}{\shortstack{Cross-Attention\\ Strategy}} 
            & \multirow{2}{*}{\shortstack{$\bar A$\\($L=10$)}}
            & \multirow{2}{*}{\shortstack{$\bar A$\\($L=20$)}}\\
            &&&&\\
            \hline
            Image &Text & Complementary & \textbf{80.55 (-0.00)} & \textbf{78.64 (-0.00)}\\
            Image &Text & Relevant & 80.34 \textcolor{red}{(-0.21)}& 78.24 \textcolor{red}{(-0.40)}\\
            Text &Image & Complementary & 79.83 \textcolor{red}{(-0.72)} & 77.57 \textcolor{red}{(-1.07)}\\
            Text &Image & Relevant & 80.05 \textcolor{red}{(-0.50)} & 77.83 \textcolor{red}{(-0.81)}\\ 
            
            \hline
        \end{tabular}        
    }
    
\end{table*}

\section{More Qualitative Results}
We provide more examples of the proposed exemplar masking on the MM-ImageNet-R dataset, in which the captions are generated by querying intstructBLIP, as shown in Figure~\ref{fig:vis_ib1},~\ref{fig:vis_ib2},~\ref{fig:vis_ib3},~\ref{fig:vis_ib4},~\ref{fig:vis_ib5},~\ref{fig:vis_ib6}. We also provide examples on the UPMC Food-101 dataset, as shown in Figure~\ref{fig:vis_f1},~\ref{fig:vis_f2},~\ref{fig:vis_f3},~\ref{fig:vis_f4}. 
For the masked images, we visualize the masked regions (i.e., $M_I \otimes x_I$) and the discarded regions (i.e., $(1-M_I) \otimes x_I$) with the corresponding mask maps. 
For the masked texts, we highlight the words related to the target objects (with the yellow color box) and the words related to the contextual information from the discarded regions (with the melon color box). 

As shown in these examples, the image masks correctly bound the class-related regions that contain the most important information. Moreover, the preserved regions are quite smaller than the discarded regions, meaning that in the image modality, there is a large proportion of redundant information that requires large storage space but is not helpful for model learning recognition.
On the other hand, the masked texts preserve both the words related to the corresponding class and the words indicating the contextual information from the discarded regions. These preserved words not only provide complementary information related to the class object but also preserve the information from the discarded image regions.

\section{Training Procedure}
Algorithm~\ref{algo:training_procedure} shows the whole training procedure of our exemplar masking framework for multimodal incremental learning. In each incremental phase, we first train the model with the available data, including new samples and exemplars. During training, we adopt multimodal data augmentation on the exemplars to replay the old knowledge effectively. After finishing the training step, we use the learned model to obtain the attention maps of the new training samples. Then we calculate the masking thresholds and obtain the resultant masks for both modalities. Finally, we apply herding algorithm~\cite{rebuffi2017icarl} to select $k$ samples as the exemplars, where the size of $k$ samples does not exceed the budget limit. 

\begin{algorithm}[t]
\For{each incremental phase}{
\KwData{Training set contains new training data $D^{\text{new}}$ and exemplars $D^{\text{exp}}$ in the $l$ incremental phase.}
    \textbf{Train the model with training set $D^{\text{new}} \cup D^{\text{exp}}$}
    \\\For{Each epoch}{
        \For{$(x_T,x_I, y)$ {\upshape in} $D^{\text{\upshape new}} \cup D^{\text{\upshape  exp}}$}{
            \If{$(x_T,x_I, y)$ {\upshape in} $D^{\text{\upshape 
 exp}}$}
            {
                Random select $x_T'$, where $y'=y$
                \\Create the augmented sample $(x_T',x_I, y)$
            }
        Update learnable parameters via $\mathcal{L}_{\text{CE}}$ in Eq. \blue{6}
        }
    }

    \textbf{Build exemplars}
    \\\For{$(x_T,x_I, y)$ {\upshape in} $D^{\text{\upshape new}}$}{
    \nonl\textbf{Mask image tokens}
    \\Calculate the attention map $A_{CLS \to I}$, and obtain the image threshold $\tau_I$ via Eq. \blue{1}
    \\Obtain image masks $M_I$ via Eq. \blue{2}
    \\ \nonl\textbf{Mask text tokens}
    \\Calculate the cross-attention map $A_{I \to T}$, and obtain the text threshold $\tau_T$ via Eq. \blue{3}
    \\Obtain text masks $M_T$ via Eq. \blue{4}
    \\ Obtain the masked samples via Eq. \blue{5}
    }
    
    \textbf{Select exemplars under the memory limit}
    \\\For{$c$ in $C^{\text{\upshape new}(l)}$}{
    Select the $k$ masked samples with the features nearest to the class mean $\mu_c$ as exemplars without exceeding the storage space.
    }
}
\caption{Training procedure of the exemplar masking framework.}
\label{algo:training_procedure}
\end{algorithm}

\section{Implementation Details}
\noindent\textbf{Inputs.}
In our experiment, we validate the proposed method on the vision and language dataset, which consists of image and text modality. For the image modality, we follow~\cite{kim2021vilt} to resize the shorter side of input images to 384 and constrain the longer side to under 640 while keeping the aspect ratio. Following \cite{dosovitskiy2020vit}, we decompose images into patches of size $32 \times 32$.
For the text modality, the text input is tokenized by the bert-base-uncased tokenizer with the maximum length of text inputs set to 128.

\noindent\textbf{Model Configurations.}
In the multimodal incremental learning framework, we have two components, including the multimodal backbone and the incremental classifier. We adopt the pre-trained multimodal transformer ViLT~\cite{kim2021vilt} as our backbone for feature extraction since it is widely used in various transformer-based methods for multimodal learning. Based on Vision Transformers~\cite{dosovitskiy2020vit}, ViLT advances to process multimodal inputs with the tokenized texts and patched images, and is pre-trained on several large vision-language datasets (e.g., MS-COCO~\cite{lin2014microsoft} and Visual Genome~\cite{krishna2017visual}) via objectives such as Image Text Matching and Masked Language Modeling.
The incremental classifier is a single linear layer that maps the features to the class prediction. As the new classes come in sequentially, in each incremental phase, we extend the classifier with additional normal-initialized parameters for the new classes.

\noindent\textbf{Model Training Details.}
In our experiments, we apply our exemplar masking framework on two learning methods, including finetuning and parameter-efficient tuning (PET) methods (i.e., SSF~\cite{lian2022ssf}). 
For the finetuning, to prevent the dramatic overriding of the weights of pre-trained ViLT backbone meanwhile learning recognition of new classes, we set the learning rate of the ViLT backbone and the classifier to $1\times 10^{-5}$ and $1\times 10^{-3}$ respectively.
For the SSF, we freeze all the parameters of the ViLT backbone and only train the learnable parameters (i.e., scales and shifts vectors for SSF) in each layer as well as the parameters of the classifier. We set the learning rate for all learnable parameters to $1\times 10^{-3}$.
We use the AdamW optimizer~\cite{loshchilov2017decoupled} in all experiments and weight decay is set to $2\times 10^{-2}$. The learning rate is warmed up for 10\% of the total training epochs and is then decreased linearly to zero. For each incremental phase, we train the models by 30 epochs.


%% file: sec/1_vis_instructBLIP.tex
\begin{figure*}[ht]
    \centering
    \Large
    \resizebox{0.95\linewidth}!{
    \begin{tabular}{rccccc}
    \centering
    &\multicolumn{2}{c}{\textrm{$x_I$}} &&\multicolumn{2}{c}{\textrm{$A_{CLS \to I}$}} \\
    &\multicolumn{2}{c}{\includegraphics[width=0.35\textwidth]{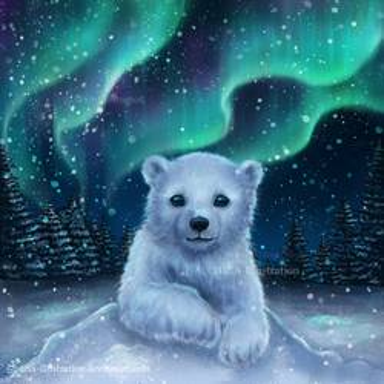}} &&\multicolumn{2}{c}{\includegraphics[width=0.35\textwidth]{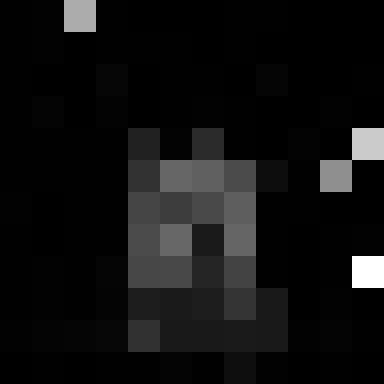}} \\
    \midrule
    &{$M_I$} & { $M_I \otimes x_I$} && {$(1-M_I)$} & {$(1-M_I) \otimes x_I$}
    \\
    &\fbox{\includegraphics[width=0.35\textwidth]{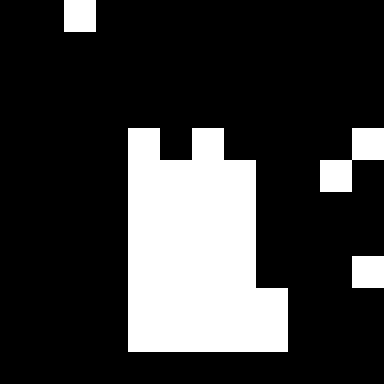}} & \includegraphics[width=0.35\textwidth]{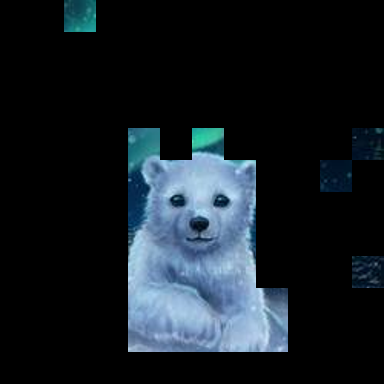}&& \fbox{\includegraphics[width=0.35\textwidth]{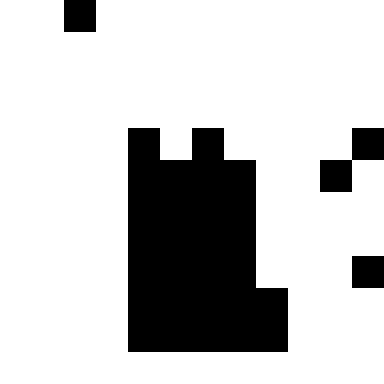}} & \includegraphics[width=0.35\textwidth]{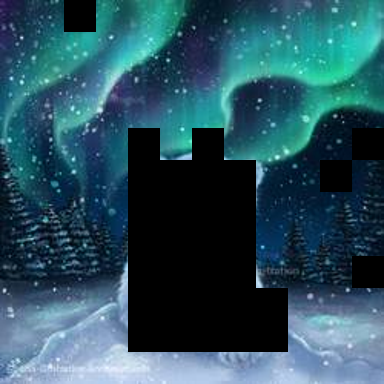}
    \\
    \midrule \midrule
    $M_T\otimes x_T$: 
    &\multicolumn{5}{p{26cm}}{[CLS] \gray{the image depicts \red{a \obj{white}} polar bear sitting \red{on} top \red{of} a \red{\bg{snow}} - covered \red{\bg{surface}} during the \red{\bg{aurora borealis}}, creating a \red{\bg{mesmerizing and serene atmosphere}}. this \red{is} a painting made by an artist \red{who} captures the beauty of the natural \red{world} in \red{her} artworks. the picture features a white polar bear sitting \red{on} a \red{\bg{snow}} - covered surface, with a vibrant and colorful northern lights backdrop illuminating the scene. the \red{\bg{white and blue}} color scheme complements the \red{\bg{aurora borealis}}, making it a stunning painting that showcases the majesty of this natural phenomenon while highlighting the beauty \red{of} the polar bear species \red{as well}.} [SEP]}\vspace{-1em}
    \end{tabular}
    }
    \caption{An example of exemplar masking on the MM-ImageNet-R (texts generated by InstructBLIP) for the class ``\textbf{ice bear}''.}
	\label{fig:vis_ib1}
\end{figure*}

\begin{figure*}[ht]
    \centering
    \Large
    \resizebox{0.95\linewidth}!{
    \begin{tabular}{rccccc}
    \centering
    &\multicolumn{2}{c}{\textrm{$x_I$}} &&\multicolumn{2}{c}{\textrm{$A_{CLS \to I}$}} \\
    &\multicolumn{2}{c}{\includegraphics[width=0.35\textwidth]{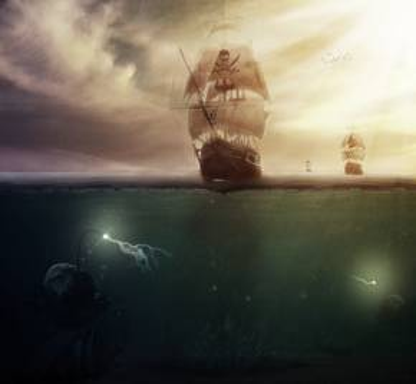}} &&\multicolumn{2}{c}{\includegraphics[width=0.35\textwidth]{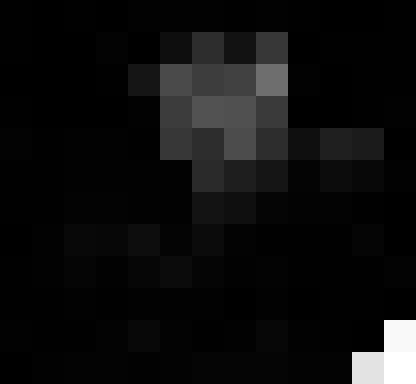}} \\
    \midrule
    &{$M_I$} & { $M_I \otimes x_I$} && {$(1-M_I)$} & {$(1-M_I) \otimes x_I$}
    \\
    &\fbox{\includegraphics[width=0.35\textwidth]{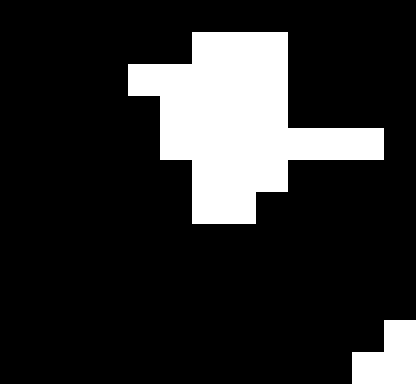}} & \includegraphics[width=0.35\textwidth]{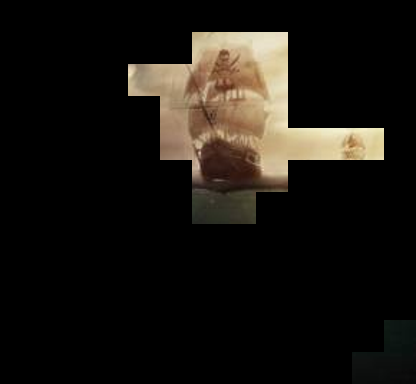}&& \fbox{\includegraphics[width=0.35\textwidth]{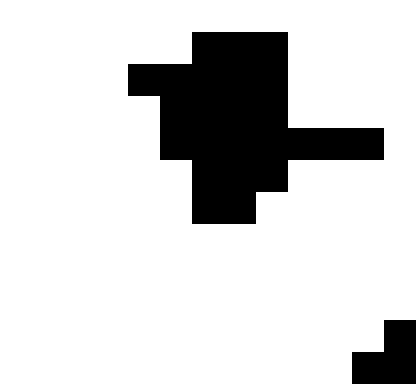}} & \includegraphics[width=0.35\textwidth]{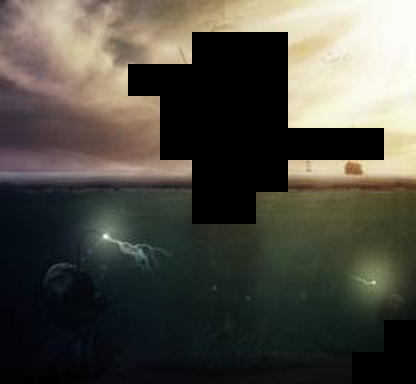}
    \\
    \midrule \midrule
    $M_T\otimes x_T$: 
    &\multicolumn{5}{p{26cm}}{[CLS] \gray{in the \red{image, there is} an \red{\obj{enormous pirate ship}} floating in \red{the \bg{ocean}} at \red{\bg{sunset}}. \red{\obj{the ship} is} positioned under a dark cloud and surrounded by various sea creatures, such as \red{\bg{squids}}, fishes, and sharks. \red{the} scene \red{oozes} a sense of \red{wonder} and fear due to \red{the \bg{ominous}} atmosphere and \red{the} presence of \red{the} \red{\bg{fearsome}} creatures. \red{the} ship, along with the vast ocean and cloudy \red{\bg{sky}}, forms a dramatic and \red{\bg{intimidating}} backdrop for \red{the} various sea creatures and adds to \red{the eerie} tone of \red{the} scene.} [SEP]}\vspace{-1em}
    \end{tabular}
    }
    \caption{An example of exemplar masking on the MM-ImageNet-R (texts generated by InstructBLIP) for the class ``\textbf{pirate ship}''.}
	\label{fig:vis_ib2}
\end{figure*}

\begin{figure*}[ht]
    \centering
    \Large
    \resizebox{0.95\linewidth}!{
    \begin{tabular}{rccccc}
    \centering
    &\multicolumn{2}{c}{\textrm{$x_I$}} &&\multicolumn{2}{c}{\textrm{$A_{CLS \to I}$}} \\
    &\multicolumn{2}{c}{\includegraphics[width=0.35\textwidth]{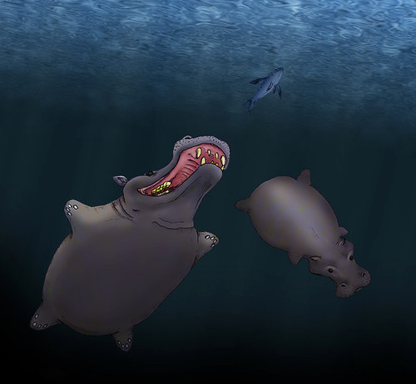}} &&\multicolumn{2}{c}{\includegraphics[width=0.35\textwidth]{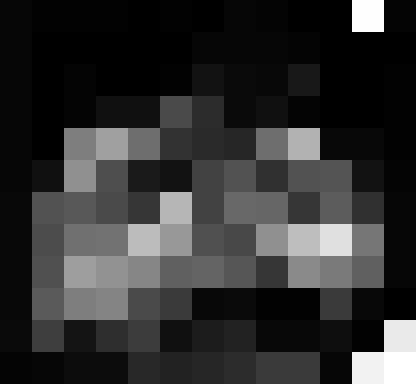}} \\
    \midrule
    &{$M_I$} & { $M_I \otimes x_I$} && {$(1-M_I)$} & {$(1-M_I) \otimes x_I$}
    \\
    &\fbox{\includegraphics[width=0.35\textwidth]{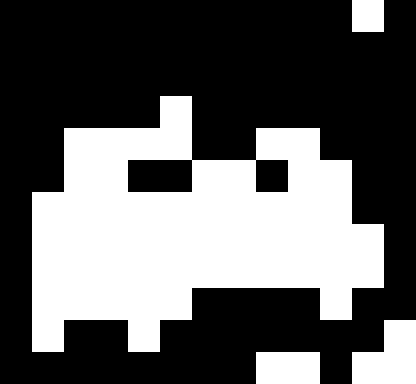}} & \includegraphics[width=0.35\textwidth]{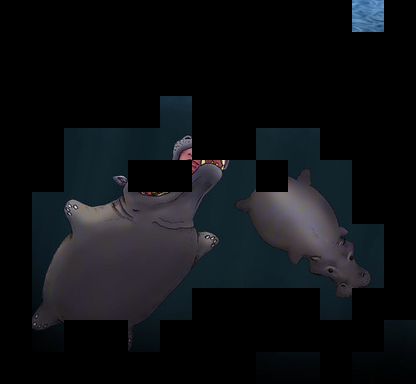}&& \fbox{\includegraphics[width=0.35\textwidth]{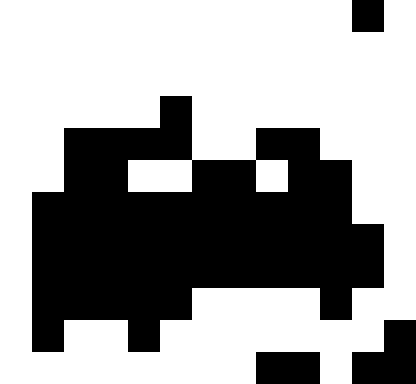}} & \includegraphics[width=0.35\textwidth]{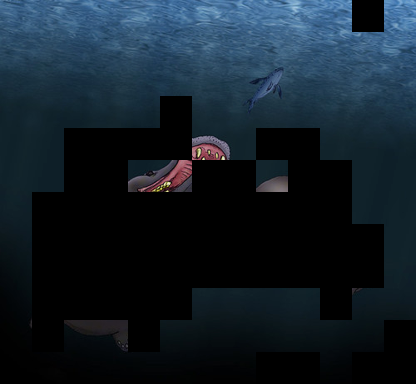}
    \\
    \midrule \midrule
    $M_T\otimes x_T$: 
    &\multicolumn{5}{p{26cm}}{[CLS] \gray{the \red{image is of} a \red{peaceful scene featuring} a \red{\bg{shark}} and \red{\obj{two hippos}}, all submerged underwater. \red{two} large \red{\obj{hippos}} \red{are present} underwater, one near the background and the other \red{nearer} to the foreground. one \red{\obj{hippo} has} a visible \red{\obj{open}} mouth, almost as if about to start swimming or perhaps looking for food. also, a \red{\bg{shark}} can be seen beneath the \red{\bg{waves}} in the foreground. the colors in the image \red{are \bg{deep}} and \red{contrasting}, with \red{\bg{neutral blue tones} emphasizing} the \red{\bg{underwater}} environment and the creatures swimming and living within it.} [SEP]}\vspace{-1em}
    \end{tabular}
    }
    \caption{An example of exemplar masking on the MM-ImageNet-R (texts generated by InstructBLIP) for the class ``\textbf{hippopotamus}''.\vspace{-1em}}
	\label{fig:vis_ib3}
\end{figure*}

\begin{figure*}[ht]
    \centering
    \Large
    \resizebox{0.95\linewidth}!{
    \begin{tabular}{rccccc}
    \centering
    &\multicolumn{2}{c}{\textrm{$x_I$}} &&\multicolumn{2}{c}{\textrm{$A_{CLS \to I}$}} \\
    &\multicolumn{2}{c}{\includegraphics[width=0.35\textwidth]{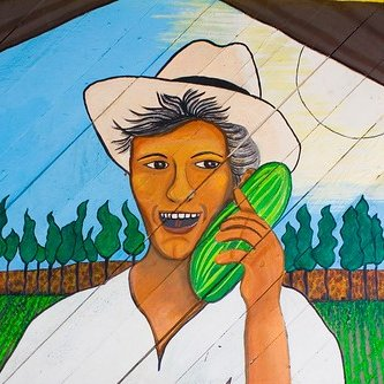}} &&\multicolumn{2}{c}{\includegraphics[width=0.35\textwidth]{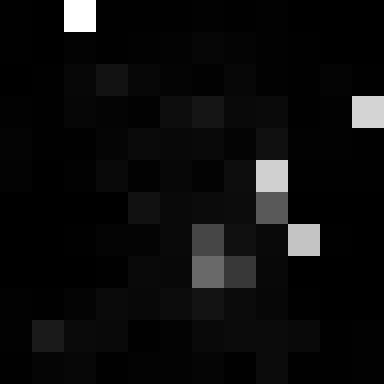}} \\
    \midrule
    &{$M_I$} & { $M_I \otimes x_I$} && {$(1-M_I)$} & {$(1-M_I) \otimes x_I$}
    \\
    &\fbox{\includegraphics[width=0.35\textwidth]{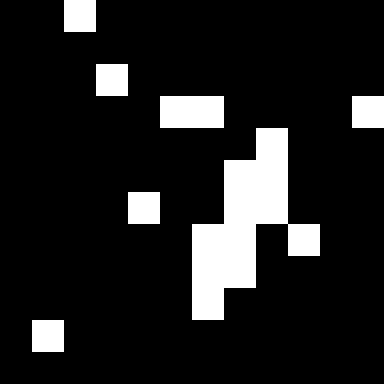}} & \includegraphics[width=0.35\textwidth]{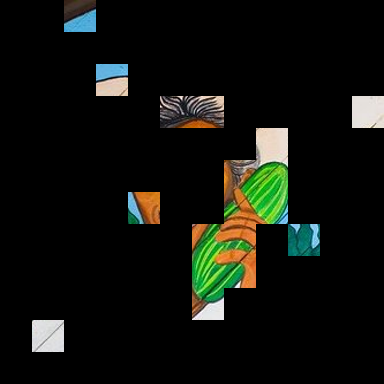}&& \fbox{\includegraphics[width=0.35\textwidth]{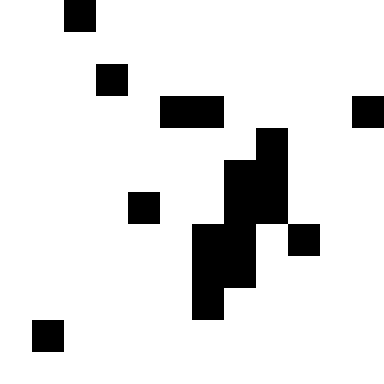}} & \includegraphics[width=0.35\textwidth]{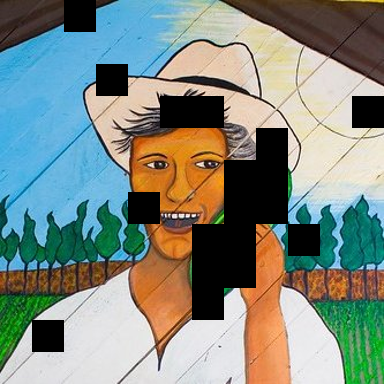}
    \\
    \midrule \midrule
    $M_T\otimes x_T$: 
    &\multicolumn{5}{p{26cm}}{[CLS] \gray{ \red{the image} features \red{a }painted \red{mural} located in \red{\bg{chiapas}}, \red{\bg{mexico}} of \red{a \textbf{farmer}} holding a melon or \red{\obj{cucumber}} in his hand while talking \red{on} a \red{phone}. the painting, which \red{is} located near a road \red{or pathway}, showcases a vivid scene of a local \red{\bg{farmer}} using communication \red{technology}, reflecting how modern life \red{is evolving} within the traditional \red{\bg{mexican farming}} setting. the painted backdrop behind the \red{farmer} has a \red{\bg{wooden}} structure \red{or \bg{fence}}, indicating the rural nature \red{of} this area and the importance of \red{\bg{agriculture}} in the local community. overall, the image captures the essence of a busy \red{\bg{farmer}}, who is balancing modern communication and traditional \red{\bg{farm}} work, \red{illustrating} the unique \red{blend}} [SEP]}\vspace{-1em}
    \end{tabular}
    }
    \caption{An example of exemplar masking on the MM-ImageNet-R (texts generated by InstructBLIP) for the class ``\textbf{cucumber}''.}
	\label{fig:vis_ib4}
\end{figure*}

\begin{figure*}[ht]
    \centering
    \Large
    \resizebox{0.95\linewidth}!{
    \begin{tabular}{rccccc}
    \centering
    &\multicolumn{2}{c}{\textrm{$x_I$}} &&\multicolumn{2}{c}{\textrm{$A_{CLS \to I}$}} \\
    &\multicolumn{2}{c}{\includegraphics[width=0.35\textwidth]{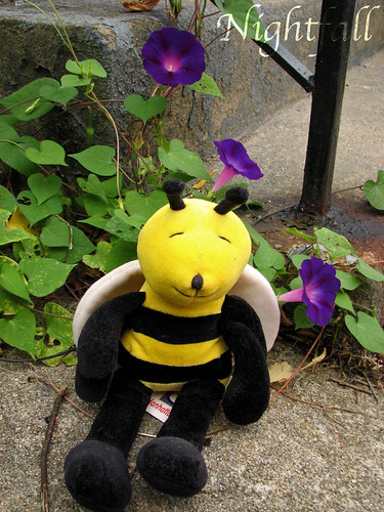}} &&\multicolumn{2}{c}{\includegraphics[width=0.35\textwidth]{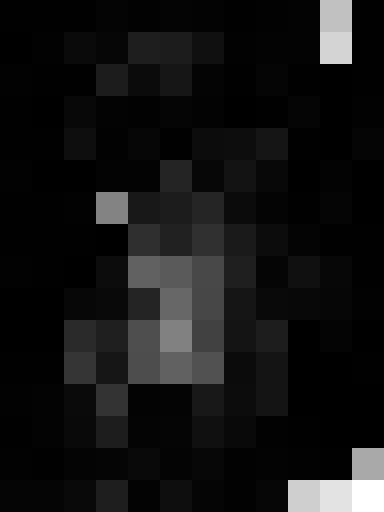}} \\
    \midrule
    &{$M_I$} & { $M_I \otimes x_I$} && {$(1-M_I)$} & {$(1-M_I) \otimes x_I$}
    \\
    &\fbox{\includegraphics[width=0.35\textwidth]{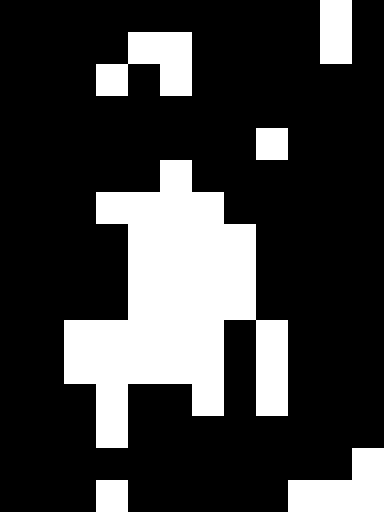}} & \includegraphics[width=0.35\textwidth]{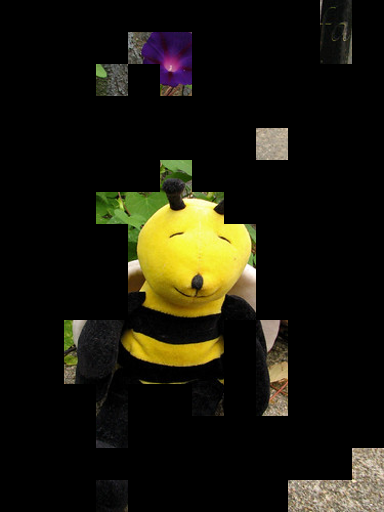}&& \fbox{\includegraphics[width=0.35\textwidth]{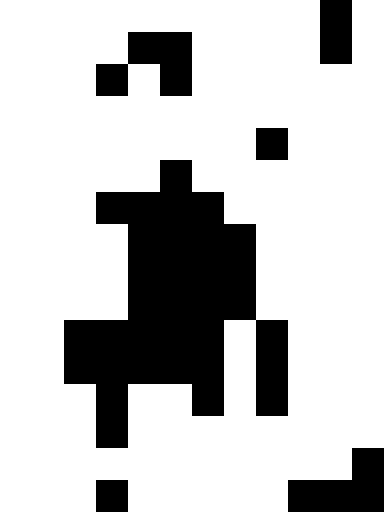}} & \includegraphics[width=0.35\textwidth]{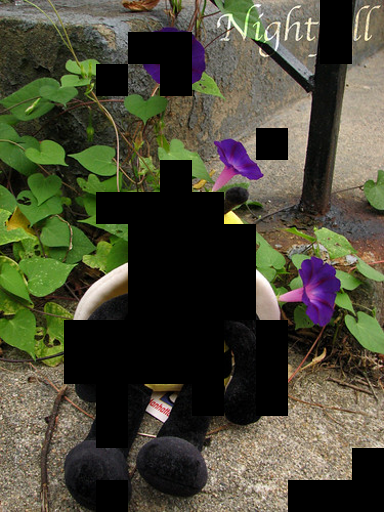}
    \\
    \midrule \midrule
    $M_T\otimes x_T$: 
    &\multicolumn{5}{p{26cm}}{[CLS] \gray{\red{the image features} a stuffed \red{\obj{bee}} sitting on the steps of a small staircase, with \red{\bg{purple flowering}} plants surrounding it. the \red{\obj{bee}} is the centerpiece of the picture and resembles a \red{cute and fun }decorative element. the flowers \red{and} plant\red{ \bg{life add some vibrancy}} to the setting. to make this scene \red{more interesting} and fun, \red{a} stuffed \red{raccoon} wearing \red{\bg{a hat} is} seen sitting by the steps, interacting with the \red{bee}.} [SEP]}\vspace{-1em}
    \end{tabular}
    }
    \caption{An example of exemplar masking on the MM-ImageNet-R (texts generated by InstructBLIP) for the class ``\textbf{bee}''.}
	\label{fig:vis_ib5}
\end{figure*}

\begin{figure*}[ht]
    \centering
    \Large
    \resizebox{0.95\linewidth}!{
    \begin{tabular}{rccccc}
    \centering
    &\multicolumn{2}{c}{\textrm{$x_I$}} &&\multicolumn{2}{c}{\textrm{$A_{CLS \to I}$}} \\
    &\multicolumn{2}{c}{\includegraphics[width=0.35\textwidth]{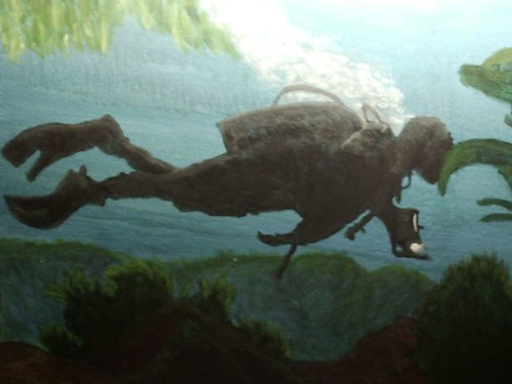}} &&\multicolumn{2}{c}{\includegraphics[width=0.35\textwidth]{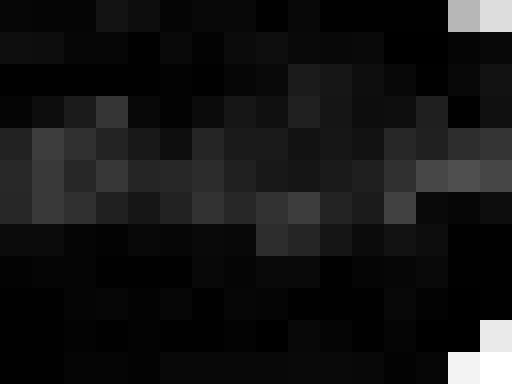}} \\
    \midrule
    &{$M_I$} & { $M_I \otimes x_I$} && {$(1-M_I)$} & {$(1-M_I) \otimes x_I$}
    \\
    &\fbox{\includegraphics[width=0.35\textwidth]{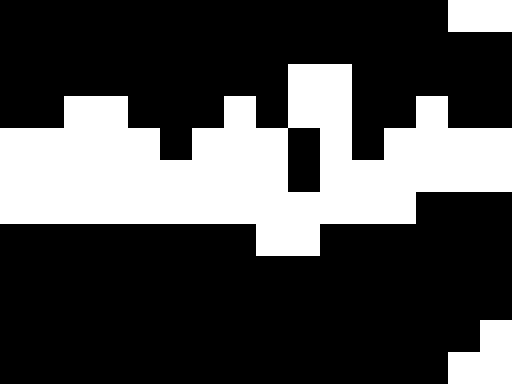}} & \includegraphics[width=0.35\textwidth]{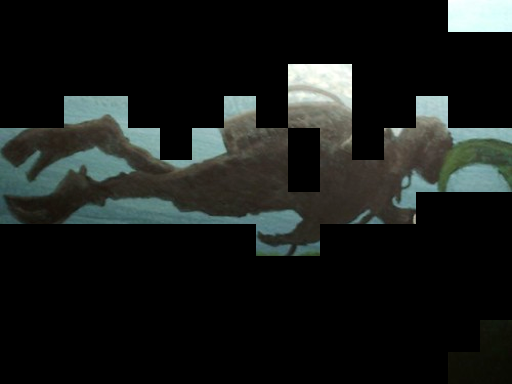}&& \fbox{\includegraphics[width=0.35\textwidth]{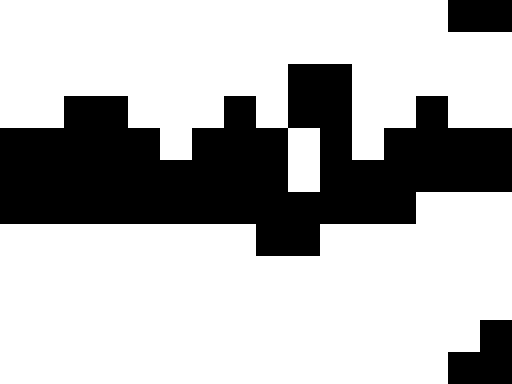}} & \includegraphics[width=0.35\textwidth]{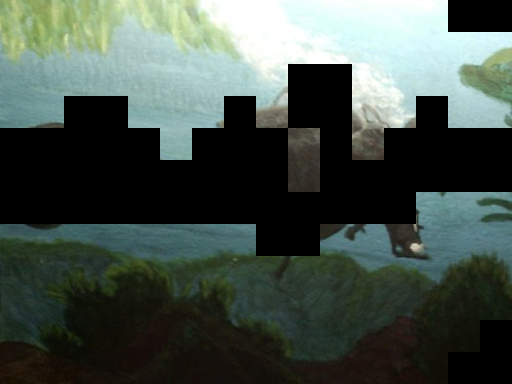}
    \\
    \midrule \midrule
    $M_T\otimes x_T$: 
    &\multicolumn{5}{p{26cm}}{[CLS] \gray{\red{the scene depicts} a \red{\obj{diver}} exploring underwater, with\red{ the focus on} the scuba gear he himself \red{is} wearing and a particular area of the underwater life. the painting displays the \red{\obj{scuba}} gear and a \red{pear}, indicating this \red{might} be a diving experience, \red{and} possibly showcased a \red{particularly} detailed and well - executed scene \red{of} an underwater experience by the artist. while the painting \red{is} mostly of the scuba gear, it's important to note that the diver can be seen \red{as} well, suggesting a focus \red{\bg{on the underwater}} environment that the diver \red{explores}.} [SEP]}\vspace{-1em}
    \end{tabular}
    }
    \caption{An example of exemplar masking on the MM-ImageNet-R (texts generated by InstructBLIP) for the class ``\textbf{scuba diver}''.}
	\label{fig:vis_ib6}
\end{figure*}

%% file: sec/1_vis_Food101.tex
\begin{figure*}[ht]
    \centering
    \Large
    \resizebox{0.95\linewidth}!{
    \begin{tabular}{rccccc}
    \centering
    &\multicolumn{2}{c}{\textrm{$x_I$}} &&\multicolumn{2}{c}{\textrm{$A_{CLS \to I}$}} \\
    &\multicolumn{2}{c}{\includegraphics[width=0.35\textwidth]{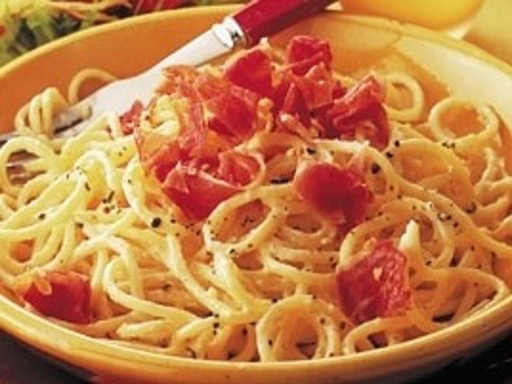}} &&\multicolumn{2}{c}{\includegraphics[width=0.35\textwidth]{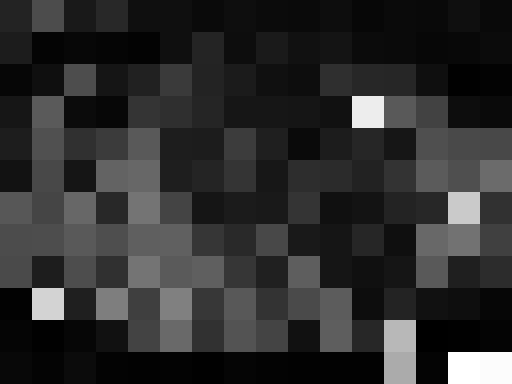}} \\
    \midrule
    &{$M_I$} & { $M_I \otimes x_I$} && {$(1-M_I)$} & {$(1-M_I) \otimes x_I$}
    \\
    &\fbox{\includegraphics[width=0.35\textwidth]{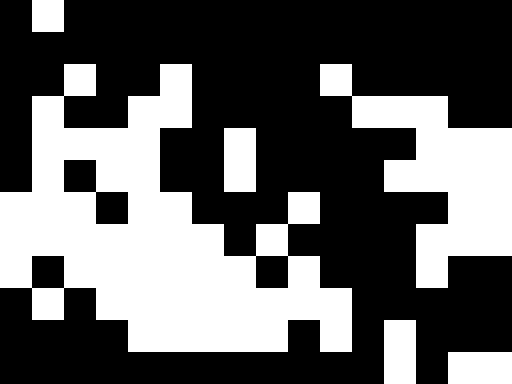}} & \includegraphics[width=0.35\textwidth]{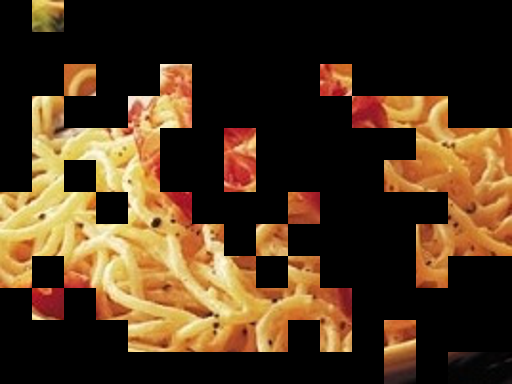}&& \fbox{\includegraphics[width=0.35\textwidth]{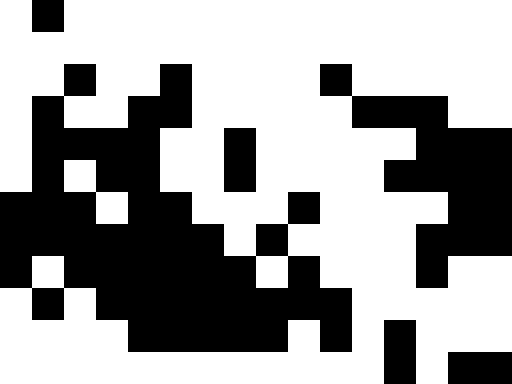}} & \includegraphics[width=0.35\textwidth]{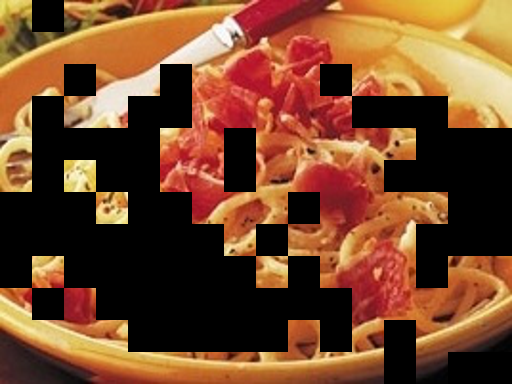}
    \\
    \midrule \midrule
    $M_T\otimes x_T$: 
    &\multicolumn{5}{p{26cm}}{[CLS] \gray{\red{\obj{spaghetti carbonara}} with roasted \red{\bg{tomato salad}} | recipes | eat well | best health} [SEP]}\vspace{-1em}
    \end{tabular}
    }
    \caption{An example of exemplar masking on the UPMC Food-101 for the class ``\textbf{spaghetti carbonara}''.}
	\label{fig:vis_f1}
\end{figure*}

\begin{figure*}[ht]
    \centering
    \Large
    \resizebox{0.95\linewidth}!{
    \begin{tabular}{rccccc}
    \centering
    &\multicolumn{2}{c}{\textrm{$x_I$}} &&\multicolumn{2}{c}{\textrm{$A_{CLS \to I}$}} \\
    &\multicolumn{2}{c}{\includegraphics[width=0.35\textwidth]{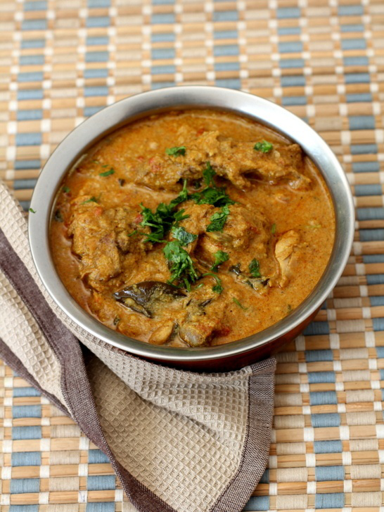}} &&\multicolumn{2}{c}{\includegraphics[width=0.35\textwidth]{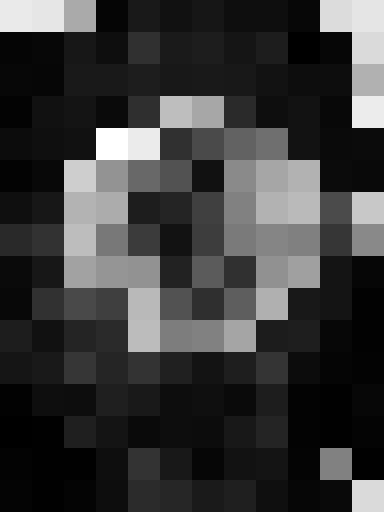}} \\
    \midrule
    &{$M_I$} & { $M_I \otimes x_I$} && {$(1-M_I)$} & {$(1-M_I) \otimes x_I$}
    \\
    &\fbox{\includegraphics[width=0.35\textwidth]{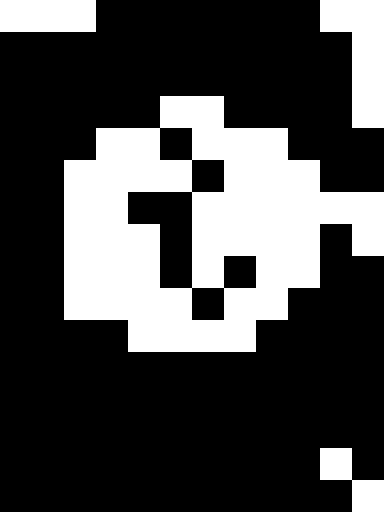}} & \includegraphics[width=0.35\textwidth]{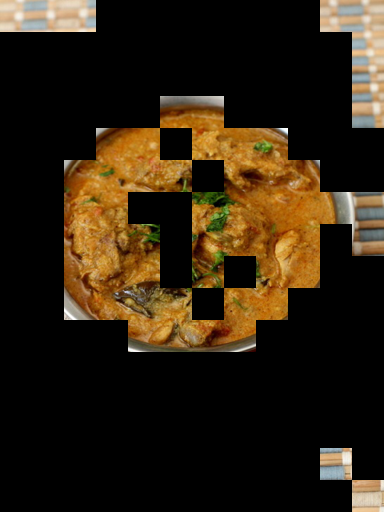}&& \fbox{\includegraphics[width=0.35\textwidth]{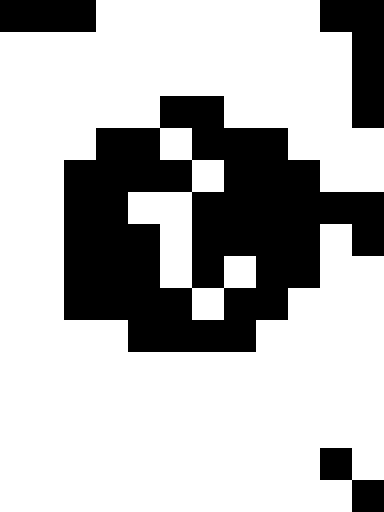}} & \includegraphics[width=0.35\textwidth]{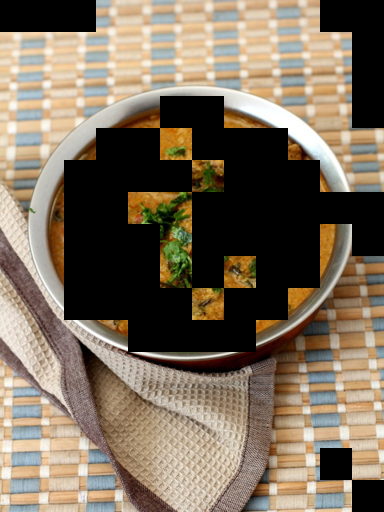}
    \\
    \midrule \midrule
    $M_T\otimes x_T$: 
    &\multicolumn{5}{p{26cm}}{[CLS] \gray{chicken salna recipe - quick chicken \red{\obj{curry}} tamil nadu style for parotta \& raquo ; all \red{recipes indian \obj{chicken} recipes indian} non - \red{\bg{vegetarian}} recipes south indian recipes} [SEP]}\vspace{-1em}
    \end{tabular}
    }
    \caption{An example of exemplar masking on the UPMC Food-101 for the class ``\textbf{chicken curry}''.}
	\label{fig:vis_f2}
\end{figure*}

\begin{figure*}[ht]
    \centering
    \Large
    \resizebox{0.95\linewidth}!{
    \begin{tabular}{rccccc}
    \centering
    &\multicolumn{2}{c}{\textrm{$x_I$}} &&\multicolumn{2}{c}{\textrm{$A_{CLS \to I}$}} \\
    &\multicolumn{2}{c}{\includegraphics[width=0.35\textwidth]{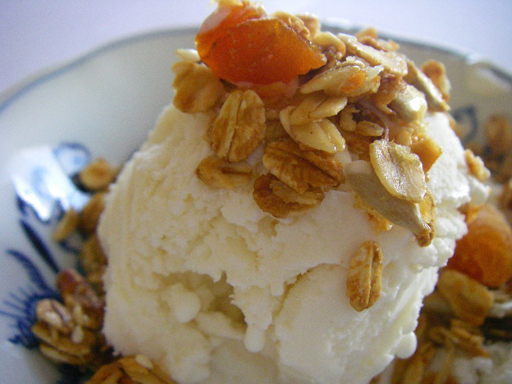}} &&\multicolumn{2}{c}{\includegraphics[width=0.35\textwidth]{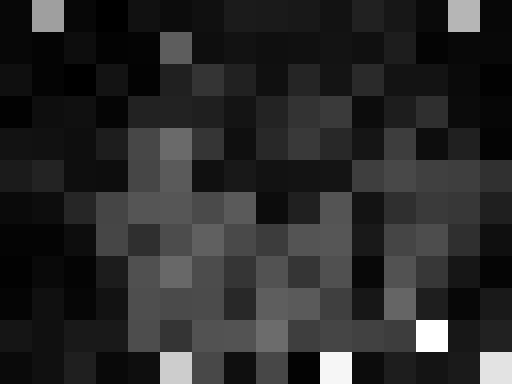}} \\
    \midrule
    &{$M_I$} & { $M_I \otimes x_I$} && {$(1-M_I)$} & {$(1-M_I) \otimes x_I$}
    \\
    &\fbox{\includegraphics[width=0.35\textwidth]{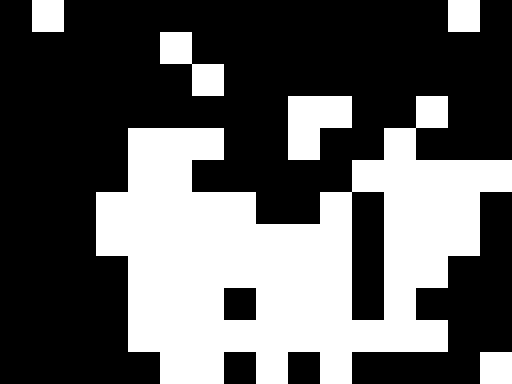}} & \includegraphics[width=0.35\textwidth]{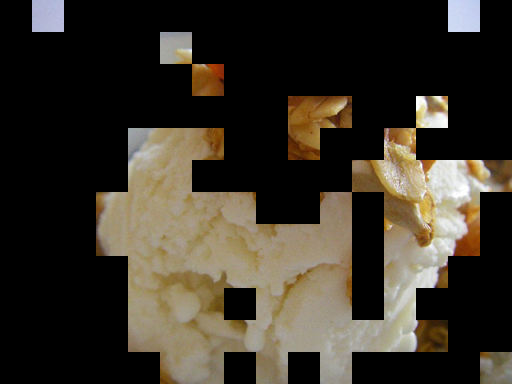}&& \fbox{\includegraphics[width=0.35\textwidth]{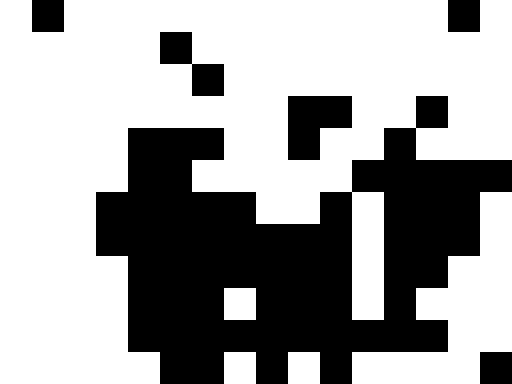}} & \includegraphics[width=0.35\textwidth]{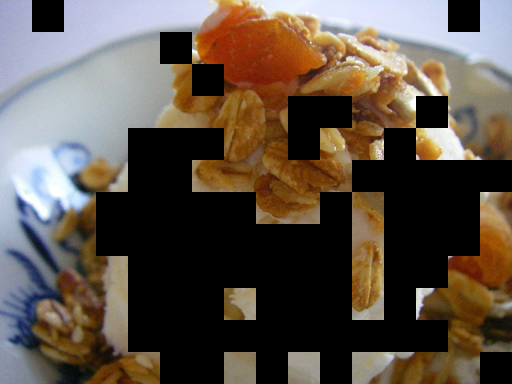}
    \\
    \midrule \midrule
    $M_T\otimes x_T$: 
    &\multicolumn{5}{p{26cm}}{[CLS] \gray{ice cream flavor of \red{the} week : \red{\bg{vanilla} \obj{frozen yogurt}} \red{with} honey \red{\bg{crunch granola}} | pink stripes } [SEP]}\vspace{-1em}
    \end{tabular}
    }
    \caption{An example of exemplar masking on the UPMC Food-101 for the class ``\textbf{frozen yogurt}''.}
	\label{fig:vis_f3}
\end{figure*}

\begin{figure*}[ht]
    \centering
    \Large
    \resizebox{0.95\linewidth}!{
    \begin{tabular}{rccccc}
    \centering
    &\multicolumn{2}{c}{\textrm{$x_I$}} &&\multicolumn{2}{c}{\textrm{$A_{CLS \to I}$}} \\
    &\multicolumn{2}{c}{\includegraphics[width=0.35\textwidth]{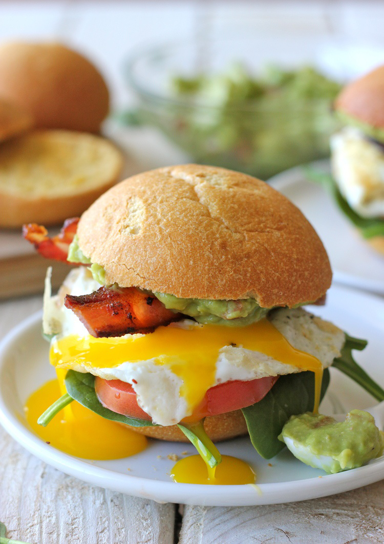}} &&\multicolumn{2}{c}{\includegraphics[width=0.35\textwidth]{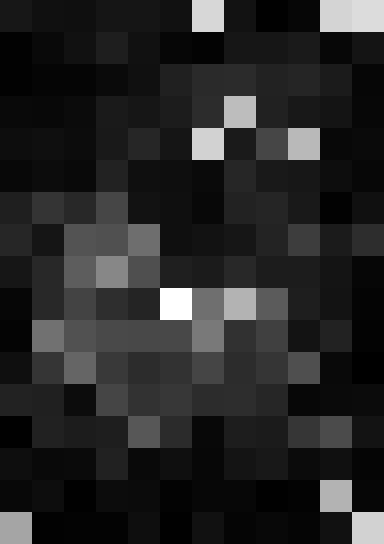}} \\
    \midrule
    &{$M_I$} & { $M_I \otimes x_I$} && {$(1-M_I)$} & {$(1-M_I) \otimes x_I$}
    \\
    &\fbox{\includegraphics[width=0.35\textwidth]{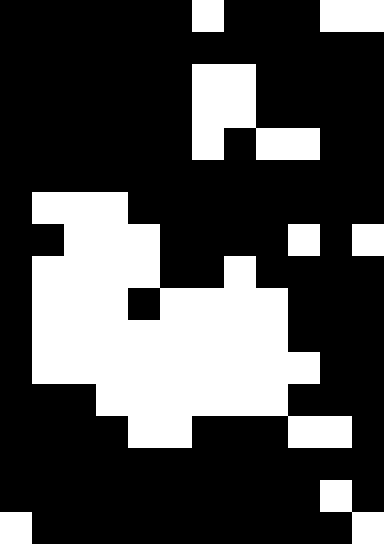}} & \includegraphics[width=0.35\textwidth]{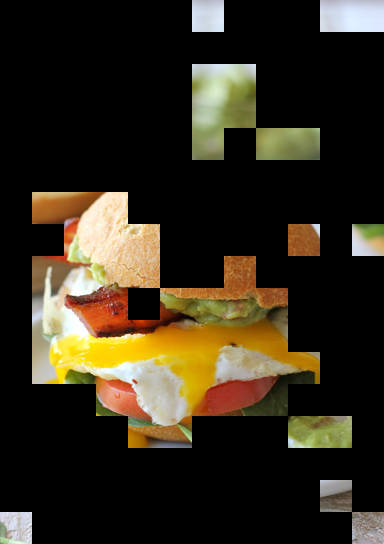}&& \fbox{\includegraphics[width=0.35\textwidth]{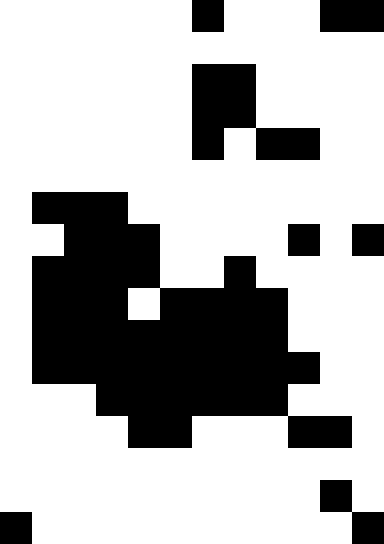}} & \includegraphics[width=0.35\textwidth]{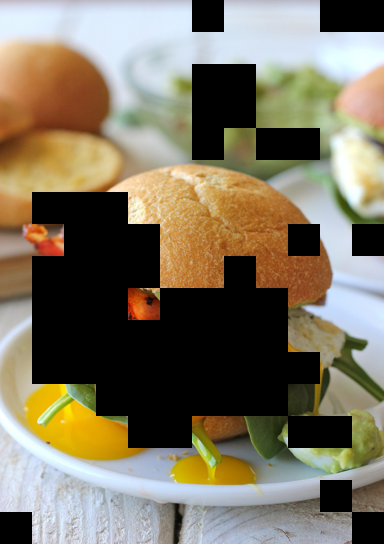}
    \\
    \midrule \midrule
    $M_T\otimes x_T$: 
    &\multicolumn{5}{p{26cm}}{[CLS] \gray{\red{\bg{avocado} \obj{club sandwich}} with \red{spicy} chipotle \red{\bg{pepper}} spread - damn delicious} [SEP]}\vspace{-1em}
    \end{tabular}
    }
    \caption{An example of exemplar masking on the UPMC Food-101 for the class ``\textbf{club sandwich}''.}
	\label{fig:vis_f4}
\end{figure*}